%% file: main.tex
\documentclass[11pt]{article}

\usepackage{acl}
\usepackage{booktabs,multirow,siunitx}
\usepackage{graphicx}
\usepackage{amsmath}
\usepackage{hyperref}
\usepackage[most]{tcolorbox}
\usepackage[table,xcdraw]{xcolor}
\usepackage{float}
\usepackage{times}
\usepackage{latexsym}
\usepackage[T1]{fontenc}
\usepackage[utf8]{inputenc}
\usepackage{microtype}
\usepackage{inconsolata}
\usepackage{amsmath, amssymb, amsfonts}
\usepackage{tabularx}
\usepackage{makecell}
\usepackage{subcaption}
\usepackage[labelformat=simple]{subcaption}

\def\method{Lang2Act}
\tcbset{
  promptbox/.style={
    enhanced,
    colback=white,
    colframe=black!15,
    boxrule=0.6pt,
    arc=3pt,
    left=8pt, right=8pt, top=8pt, bottom=8pt,
    title style={font=\bfseries},
    fontupper=\normalsize\rmfamily,
  },
}

\title{\method{}: Fine-Grained Visual Reasoning through Self-Emergent Linguistic Toolchains}

\author{
    Yuqi Xiong$^{1}$\thanks{ \ \ indicates equal contribution.}, 
    Chunyi Peng$^{1}$\footnotemark[1], 
    Zhipeng Xu$^{1}$, 
    Zhenghao Liu$^{1}$\thanks{ \ \ indicates corresponding author.}, \\ 
    \textbf{Zulong Chen$^{3}$, Yukun Yan$^{2}$, Shuo Wang$^{2}$, Yu Gu$^{1}$ and Ge Yu$^{1}$} \\
    $^1$School of Computer Science and Engineering, Northeastern University, Shenyang, China \\
    $^2$Department of Computer Science and Technology,  Tsinghua University, Beijing, China \\
    $^3$Alibaba Group, Hangzhou, China \\
}
\begin{document}
\maketitle

\input{section/0_abstract}
\input{section/1_introduction}

\input{section/2_relatedwork}

\input{section/3_method}

\input{section/4_experiment}
\input{section/5_result}
\input{section/6_conclusion}
\input{section/7_acknowledge}

% \bibliographystyle{acl_natbib}
\bibliography{citation}

\clearpage
\input{section/8_appendix}

\end{document}

%% file: section/0_abstract.tex
\begin{abstract}
Visual Retrieval-Augmented Generation (VRAG) enhances Vision-Language Models (VLMs) by incorporating external visual documents to address a given query. Existing VRAG frameworks usually depend on rigid, pre-defined external tools to extend the perceptual capabilities of VLMs, typically by explicitly separating visual perception from subsequent reasoning processes. However, this decoupled design can lead to unnecessary loss of visual information, particularly when image-based operations such as cropping are applied. In this paper, we propose \textbf{\method{}}, which enables fine-grained visual perception and reasoning through self-emergent linguistic toolchains. Rather than invoking fixed external engines, \method{} collects self-emergent actions as linguistic tools and leverages them to enhance the visual perception capabilities of VLMs. To support this mechanism, we design a two-stage Reinforcement Learning (RL)-based training framework. Specifically, the first stage optimizes VLMs to self-explore high-quality actions for constructing a reusable linguistic toolbox, and the second stage further optimizes VLMs to exploit these linguistic tools for downstream reasoning effectively. Experimental results demonstrate the effectiveness of \method{} in substantially enhancing the visual perception capabilities of VLMs, achieving performance improvements of over 4\%. All code and data are available at \url{https://github.com/NEUIR/Lang2Act}.
\end{abstract}

%% file: section/1_introduction.tex
\section{Introduction}
\input{figure/intro}
Retrieval-Augmented Generation (RAG) has been established as a foundational framework for enhancing Large Language Models (LLMs)~\citep{lewis2020retrieval, liu2025knowledge} by retrieving query-related text documents and then feeding these documents as contextual knowledge to support generation~\cite{ram2023context}. To extend the benefits of RAG to visual documents, existing methods~\citep{yu2024visrag, faysse2024colpali} typically adopt an end-to-end Visual Retrieval-Augmented Generation (VRAG) modeling paradigm. This design avoids the error propagation introduced by text-based RAG pipelines that rely on optical character recognition, instead directly leveraging the strong visual understanding capabilities of Vision-Language Models (VLMs). Specifically, VRAG models treat entire document-page snapshots as retrieval units and generate answers directly based on the retrieved visual pages. Such systems effectively preserve global visual context and layout structures, which are often lost in text-only representations of visually rich documents.

To further empower VLMs in handling complex queries, some works have focused on exploiting the reasoning capabilities of VLMs to produce more accurate answers grounded in retrieved document pages. Some of them~\citep{wu2025mmsearch, peng2025learning, sun2025visrag} employ Reinforcement Learning (RL)~\citep{shao2024deepseekmath, yu2025dapo} to optimize VLMs for generating more effective Chain-of-Thought (CoT)~\citep{wei2022chain} reasoning to derive more accurate answers. This enables VLMs to more effectively select query-relevant document pages and extract critical evidence from visual documents. By incentivizing VLMs~\citep{bai2025qwen2,yao2024minicpm,peng2025learning} to plan search steps and integrate information across multiple sources autonomously, these approaches facilitate more effective retrieval and reasoning behaviors, thereby improving the collection and utilization of relevant external knowledge.

Instead of focusing on the denoising and utilization of retrieved documents, recent research~\citep{wang2025vrag, wang2025pixel} aims to empower VLMs with finer-grained perceptual capabilities through interaction with explicit image tools. By leveraging external APIs to actively crop, zoom, or select specific image regions, these models can selectively attend to subtle visual details that are often overlooked by holistic processing. Through this simulation of active observation, tool-enhanced VLMs~\citep{wang2025vrag, wang2025pixel} seek to bridge the gap between the intrinsic reasoning abilities of VLMs and low-level pixel inspection. However, such systems typically rely on rigid, pre-defined image tools that explicitly decouple visual perception from logical reasoning, constraining models to fixed mechanical operations and potentially leading to the loss of critical visual information during tool application.

To bridge this gap and unify perception with reasoning, we propose \method{}, a framework that enables fine-grained visual reasoning via self-emergent linguistic toolchains. As shown in Figure~\ref{fig:intro}, unlike traditional methods that rely on detached external engines, \method{} curates a set of linguistic tools to enhance visual perception, such as \texttt{read\_text\_elem}, and implicitly internalizes visual operations through autoregressive generation. When a linguistic tool is decoded, it encourages the VLM to attend to image regions relevant to the tool instruction, thereby enabling more fine-grained visual perception. To optimize how VLMs curate and utilize linguistic tools, we adopt a two-stage RL-based optimization mechanism~\citep{yu2025dapo}, in which the model first discovers effective visual actions for problem solving through self-exploration, and subsequently strengthens the visual perception capability of VLMs by leveraging the curated linguistic toolbox.

Experimental results on multiple visual question answering benchmarks demonstrate the effectiveness of \method{}, which outperforms all baselines by more than 4\%. Further analysis shows that, by leveraging intrinsic linguistic toolchains, \method{} not only facilitates accurate localization of the ground-truth image regions, but also achieves higher answer accuracy when attending to golden regions. These results indicate that \method{} enhances visual perception, more effectively exploits visual evidence, and alleviates hallucination. Notably, \method{} mitigates unexpected information loss typically introduced by image tools, relying instead on linguistic tools.

%% file: figure/intro.tex
% \begin{figure}[t!]
%     \includegraphics[width=\linewidth]{image/intro.png}
%     \caption{Illustration of Our Reasoning Compression Through Stepwise Trials (ReCUT) Model.}
%     \label{fig:intro}
% \end{figure}
\begin{figure}[t!]
    \centering
    \includegraphics[width=\linewidth]{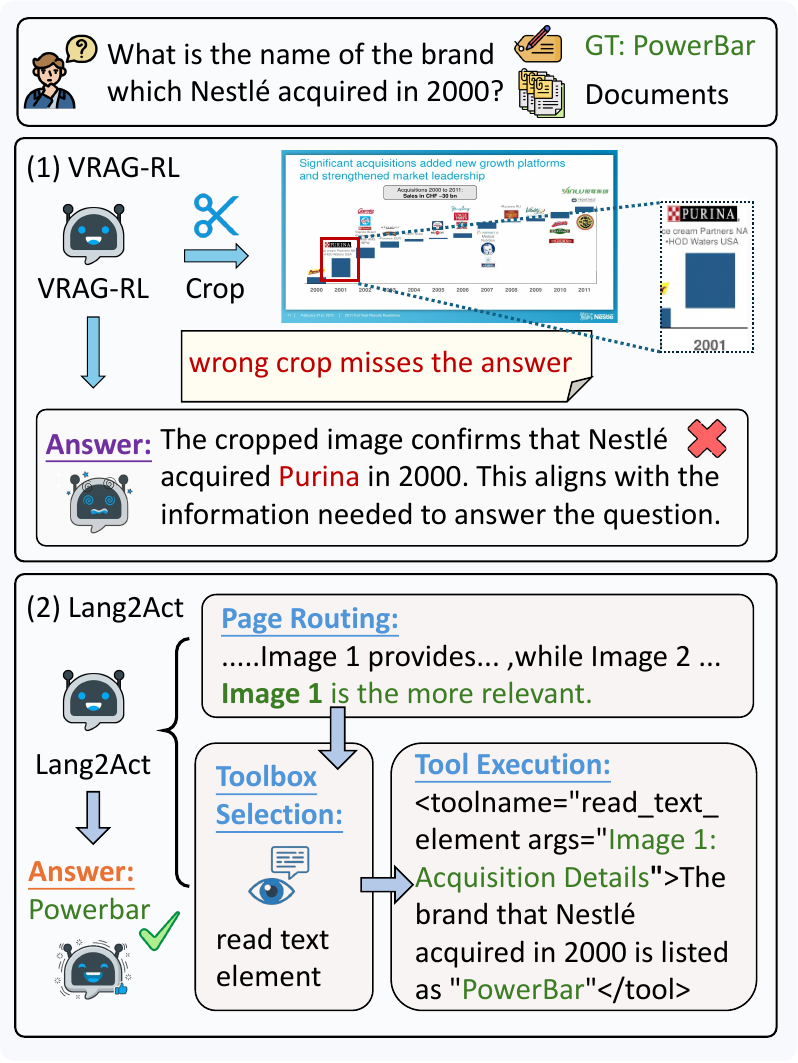}
    \caption{{Comparison between VRAG-RL and the \method{} framework.}}
    \label{fig:intro}
\end{figure}

%% file: section/2_relatedwork.tex
\section{Related Work}

Retrieval-Augmented Generation (RAG) has been established as a foundational framework for enhancing Large Language Models (LLMs) by grounding generation in external knowledge~\citep{lewis2020retrieval, gu2021unidoc, liu2025knowledge}. While effective for plain text, traditional RAG often discards critical layout cues and spatial structures when processing visually rich documents~\citep{zhang2025ocr}. To address this, Visual Retrieval-Augmented Generation (VRAG)~\citep{suri2024visdom, cho2024m3docrag} has emerged as a significant advancement. By utilizing whole document-page snapshots as retrieval units, systems like ColPali~\citep{faysse2024colpali} and VisRAG~\citep{yu2024visrag} effectively preserve the global visual context. However, despite retaining layout information, these methods typically rely on implicit holistic processing, mapping high-dimensional visual features directly to answers without the resolution required to inspect fine-grained visual details.

To further empower RAG models in handling complex and knowledge-intensive queries, recent works leverage Reinforcement Learning (RL)~\citep{shao2024deepseekmath,yu2025dapo} to enhance VLMs for both retrieval-augmented reasoning and search planning. For intrinsic reasoning, R1-Onevision~\citep{yang2025r1}, Vision-R1~\citep{huang2025vision}, and ThinkLite-VL~\citep{wang2025sota} employ RL to reward Chain-of-Thought trajectories that can accurately deduce the golden answers. For information seeking, MM-Search-R1~\citep{wu2025mmsearch} and R1-Router~\citep{peng2025learning} learn to autonomously conduct search planning, while EVisRAG~\citep{sun2025visrag} guides VLMs in visual evidence extraction and integration. Despite their success in accurate visual evidence exploitation, these methods fail to consider the role of RL in capturing richer visual cues for finer-grained visual understanding.

To enable finer-grained visual perception, recent works have augmented VLMs~\citep{bai2025qwen2, yao2024minicpm} with external visual tools. For example, Pixel-Reasoner~\citep{wang2025pixel} enables pixel-level inspection by introducing explicit operations, such as zoom and select, to attend to specific image coordinates. Likewise, VRAG-RL~\citep{wang2025vrag} formulates the retrieval–reasoning loop as a sequential decision process, allowing an agent to iteratively crop ambiguous regions for more detailed examination. Although these methods achieve improvements in image perception, they rely on rigid, tool-based interfaces, leading to fragmented reasoning processes and inevitable inference latency. More importantly, discrete visual operations can cause irreversible information loss due to imperfect cropping or suboptimal region selection.

%% file: section/3_method.tex
\section{Methodology}
\label{sec:method}
\input{figure/main}
This section presents \method{}, a unified framework for fine-grained visual reasoning with Vision-Language Models (VLMs),
as illustrated in Fig.~\ref{fig:main}. Given a question $q$ and retrieved document pages $\mathcal{P} = \{p_1,\dots,p_K\}$, the objective of the VLM is to answer the query $q$ based on retrieved documents $\mathcal{P}$.
We first introduce a linguistic tool curation process that extracts tools emerging from model reasoning trajectories (Sec.~\ref{method:toolcreate}). Building upon this curated toolset, \method{} continuously optimizes the visual understanding capability of VLMs by incorporating linguistic tools as prompts during inference and training (Sec.~\ref{method:framework}).

\subsection{Enhancing Visual Reasoning Trajectories for Linguistic Tool Curation}
\label{method:toolcreate}
To enable VLMs to perform fine-grained and interpretable visual reasoning, \method{} first optimizes VLMs to generate higher-quality visual reasoning trajectories, thereby encouraging the emergence of more effective visual understanding actions. The linguistic tools distilled from successful reasoning trajectories are then collected to form the linguistic tool pool $\mathcal{T}_{\text{pool}}$.

\textbf{Linguistic Action Emergence via RL Training.}
Existing visual understanding approaches~\cite{sun2025visrag} typically adopt Chain-of-Thought (CoT) prompting to facilitate visual reasoning in VLMs, including explicit visual actions such as ``read the title, list related information''. To collect higher-quality visual actions from reasoning trajectories, we train the VLM ($\pi_\theta$) using the Decoupled Clip and Dynamic sampling Policy Optimization (DAPO) method~\cite{yu2025dapo} to encourage self-exploration of reasoning trajectories and retain these trajectories that can lead to correct answers.

For each training sample $(q,\mathcal{P},a^*)$, where $a^*$ denotes the ground-truth answer to the question $q$, the VLM policy \(\pi_\theta\) generates a complete linguistic reasoning trajectory $\tau$ along with a final answer $a$:
\begin{equation}\small
    (\tau,a) \sim \pi_\theta(\cdot \mid q, \mathcal{P}).
\end{equation}
The model parameters \(\theta\) are then optimized to maximize the expected answer reward:
\begin{equation}\small\label{eq:dapo_1}
    \max_{\theta}\  \mathbb{E}_{\tau \sim \pi_{\theta}(\cdot|q, \mathcal{P})} [r_\text{ans}(\tau, a)],
\end{equation}
where the answer reward $r_{\text{ans}}(\tau, a) \in \{0, 1\}$ is provided by an automatic reward model that evaluates the predicted answer against the reference answer.

\textbf{Tool Curation.} 
To organize and summarize the self-explored linguistic actions, we process the reasoning trajectories sequentially while maintaining a global tool pool $\mathcal{T}_{\text{pool}}$. These linguistic tools that appear most frequently are retained in $\mathcal{T}_{\text{box}}$, and each tool $t$ in $\mathcal{T}_{\text{pool}}$ is represented as a standardized triplet consisting of a tool name, a textual description, and a parameter specification.

After obtaining the optimized exploration policy $\pi_{\theta^*}$, we sample reasoning trajectories for all training instances $(q_i, \mathcal{P}_i, a_i^*)$ in the dataset.
Specifically, for each instance, we sample a trajectory:
\begin{equation}\small
    (\tau_i,a_i) \sim \pi_{\theta^*}(\cdot \mid q_i, \mathcal{P}_i),
\end{equation}
and denote the resulting trajectory set as $\{\tau_i\}_{i=1}^{n}$.
Each trajectory $\tau_i$ contains a sequence of self-explored linguistic actions executed over the retrieved document pages.
When processing a trajectory $\tau_i$, we use the optimized VLM $\pi_{\theta^*}$ to abstract these actions into a set of utilized tools $S_i$:
% \begin{equation}\small S_i = \operatorname{Extract}(\tau_i, \mathcal{T}{\text{pool}}^{(i-1)}; \pi{\theta^*}), 
% \end{equation}
\begin{equation}\small S_i \sim \pi_{\theta^*}(\tau_i, \mathcal{T}_{\text{pool}}^{(i-1)}).
\end{equation}
The $\pi_{\theta^*}$ parses the actions in $\tau_i$ and aligns them to maintain a consistent format with the existing tool set $\mathcal{T}_{\text{pool}}^{(i-1)}$. And then the tool pool is updated using the produced tool set $S_i$:
\begin{equation}\small
    \mathcal{T}_{\text{pool}}^{(i)} = \mathcal{T}_{\text{pool}}^{(i-1)} \cup S_i,
\end{equation}
where $\mathcal{T}_{\text{pool}}^{(i)}$ denotes the tool pool after processing trajectory $\tau_i$, and the pool is initialized as $\mathcal{T}_{\text{pool}}^{(0)} = \emptyset$.

% After processing all $n$ trajectories, we select the top-$K$ most frequently used tools to construct the linguistic toolbox:
% \begin{equation}\small
% \mathcal{T}_{\text{box}}
% = \operatorname{Top}_K \big( \{ (t_1, f(t_1)), \dots, (t_m, f(t_m)) \} \big).
% \end{equation}
% Here, $t_j \in \mathcal{T}_{\text{pool}}^{(n)}$ and the ranking is determined by the usage frequency $f(t_j)$. The usage frequency of each tool $t_j$ is computed as:
% \begin{equation}\small
% f(t_j) = \sum_{i=1}^{n} \mathbf{1}[t_j \in S_i].
% \end{equation}
After processing all $n$ trajectories, we get the final tool set $\mathcal{T}_{\text{pool}}^{(n)}$ and select the top-$K$ most frequently used tools from the set to construct the linguistic toolbox:
\begin{equation}\small
\mathcal{T}_{\text{box}}
= \operatorname{Top}_K \big( \{ (t_1, f(t_1)), \dots, (t_m, f(t_m)) \} \big),
\end{equation}
where $t_m \in \mathcal{T}_{\text{pool}}^{(n)}$ and the ranking is determined by the usage frequency $f(t_m)$, computed as:
\begin{equation}\small
f(t_m) = \sum_{i=1}^{n} \mathbf{1}[t_m \in S_i].
\end{equation}

\subsection{Optimizing Vision-Language Models through Linguistic Tool-Based Prompting}
\label{method:framework}
With the curated linguistic toolbox $\mathcal{T}_{\text{box}}$ obtained in Sec.~\ref{method:toolcreate}, we further optimize the VLM parameters $\theta^*$ via reinforcement learning by augmenting the optimized VLM with linguistic tools, thereby fully leveraging the potential of these linguistic tools in visual reasoning.

\textbf{Linguistic Tool Grounded Visual Reasoning.} We internalize visual perception tools as verbalized functions within the curated toolbox $\mathcal{T}_{\text{box}}$, enabling the VLM to autoregressively generate a structured linguistic toolchain $z$ together with the final answer $a$ in a unified generation process:
\begin{equation}\small
(z, a) \sim \pi_{\theta^*}(\cdot \mid q, \mathcal{P}, \mathcal{T}_{\text{box}}).
\end{equation}
To facilitate grounded and interpretable tool usage, the linguistic toolchain $z$ is explicitly organized into two consecutive components:
\begin{equation}\small
z = \{z_{\text{route}},z_{\text{exec}}\},
\end{equation}
where $z_{\text{route}}$ corresponds to a visual routing stage that performs the page selection over the retrieved pages $\mathcal{P}$ and identifies a subset of relevant pages, denoted as $\mathcal{P}_{\text{sub}} \subseteq \mathcal{P}$, for subsequent reasoning.
Conditioned on the selected subset $\mathcal{P}_{\text{sub}}$, the second reasoning component $z_{\text{exec}}$ represents a linguistic tool execution stage that generates a sequence of language-level tool actions along with their corresponding execution observation:
\begin{equation}\small
z_{\text{exec}} = \{ (u_1, o_1), (u_2, o_2), \dots, (u_l, o_l) \},
\end{equation}
where $u$ denotes an implicit invocation of a linguistic tool from $\mathcal{T}_{\text{box}}$, implemented via autoregressive decoding. This invocation is applied to guide VLMs to focus on specific regions of the target document in $\mathcal{P}_{\text{sub}}$ encouraging more targeted visual evidence extraction. The corresponding execution observation $o$ is generated internally by the model through language-driven reasoning over the target pages.
These execution observations serve as fine-grained intermediate evidence supporting the final answer generation.

\textbf{Tool-Enhanced Reasoning Optimization.} To synergistically optimize the ability of $\pi_{\theta^*}$ to utilize the curated tools, execute them intrinsically, and integrate visual information, we also employ DAPO~\citep{yu2025dapo} by maximizing the expected reward:
\begin{equation}\small
\max_{\theta}\ \mathbb{E}_{(z,a)\sim \pi_{\theta^*}(\cdot \mid q,\mathcal{P},\mathcal{T}_{\text{box}})} \big[ R(z, a) \big],
\end{equation}
where the reward $R(z, a)$ is defined as a weighted combination to balance final answer correctness and intrinsic toolchain validity:
\begin{equation}\small\label{eq:reward}
R(z, a) = \alpha \cdot r_{\text{ans}}(z, a) + \beta \cdot r_{\text{tool}}(z),
\end{equation}
where both $\alpha$ and $\beta$ are hyperparameters. The answer reward follows Eq.~\ref{eq:dapo_1}, and the reward $r_{\text{tool}}(z)$ is a tool-usage regularization term that encourages linguistic tool usage to occur within the tool execution stage $z_{\text{exec}}$:
\begin{equation}\small
r_{\text{tool}}(z) = \mathbb{I}_{\text{valid}}(z \mid \mathcal{T}_{\text{box}}),
\end{equation}
where $\mathbb{I}_{\text{valid}}(z) = 1$ if the linguistic toolchain $z$ satisfies the prescribed structural constraints on tool usage, and $0$ otherwise.

%% file: figure/main.tex
% \begin{figure*}[t!]
%     \includegraphics[width=\textwidth]{image/overall.pdf}
%     \caption{The Overview of Our Reasoning Compression Through Stepwise Trials (ReCUT) Model.}
%     \label{fig:main}
% \end{figure*}

\begin{figure*}[t!]
    \includegraphics[width=\textwidth]{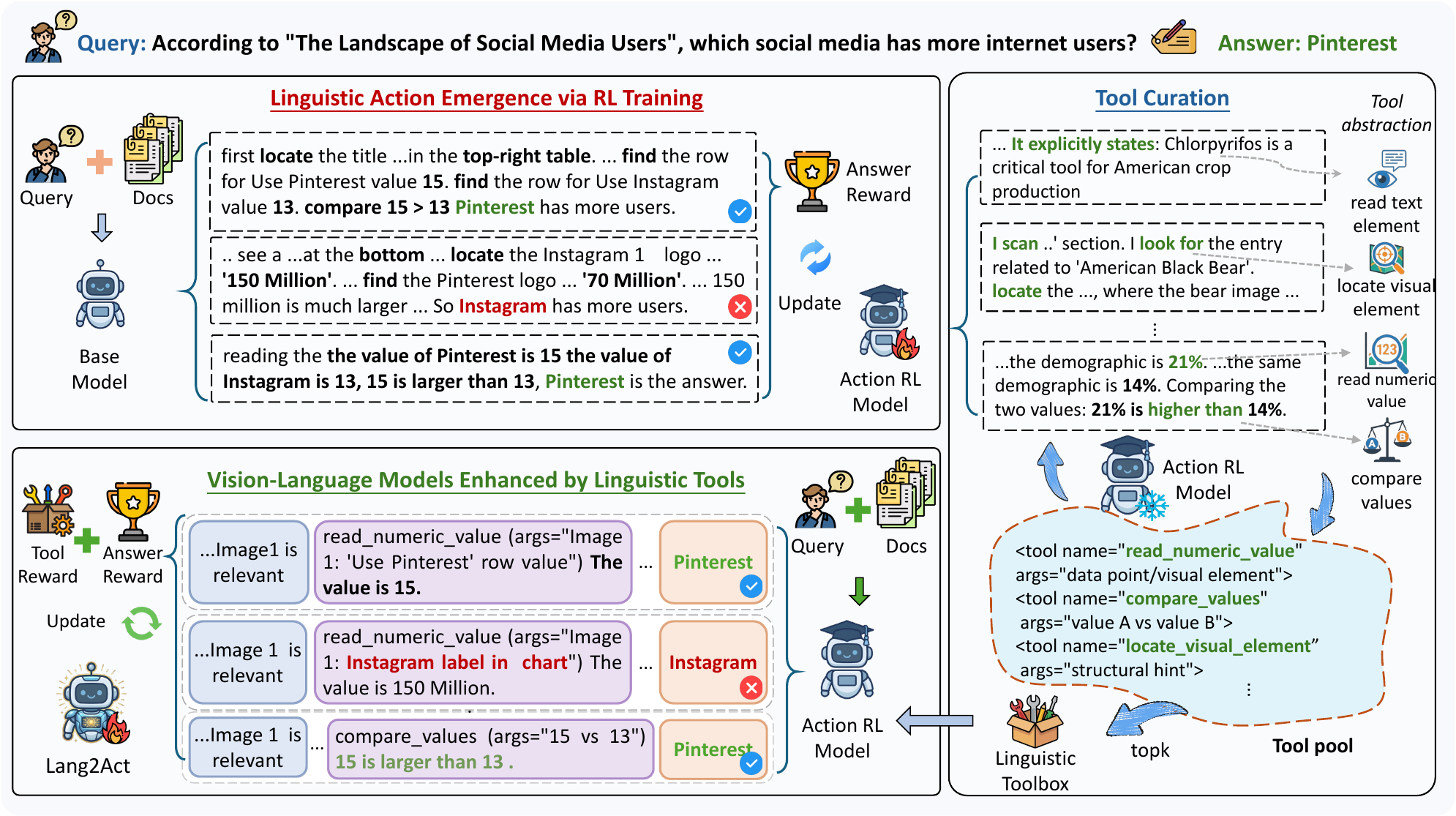}
    \caption{The overview architecture of \method{}.}
    \label{fig:main}
\end{figure*}

%% file: section/4_experiment.tex
\section{Experimental Methodology}
This section describes the dataset, evaluation metrics, baselines, and implementation details.

\textbf{Datasets.}
We evaluate \method{} on three document visual question answering
benchmarks: SlideVQA~\citep{tanaka2023slidevqa},
ViDoSeek~\citep{wang2025vidorag}, and
MMLongBench-Doc~\citep{ma2024mmlongbench}, which respectively cover
presentation slides, visually rich documents, and long-context multimodal
documents.
Additional dataset statistics and benchmark details are provided in
Appendix~\ref{app:datasets}.

\textbf{Evaluation Metrics.}
Following~\citet{wang2025vrag}, we adopt the LLM-as-judge method and calculate the accuracy as the primary evaluation metric.
Specifically, we employ Qwen2.5-72B-Instruct~\citep{bai2025qwen2} as an automatic evaluator to compare the predictions of models with the corresponding ground-truth answers.

\textbf{Baselines.}
To comprehensively validate the effectiveness of \method{}, we compare our method with four categories of approaches: \texttt{Prompting Methods}, Vision Language Reasoning Models (\texttt{VLRMs}), Multimodal Retrieval-Augmented Generation models (\texttt{MRAGs}), and \texttt{Tool-Enhanced VLMs}.

First, the \texttt{Prompting Methods} category includes Vanilla model~\citep{bai2025qwen2}, TOT~\cite{yao2023tree}, and GOT~\cite{besta2024graph}, which serve to evaluate the native reasoning capabilities of the backbone model under different chain-of-thought prompting strategies~\cite{wei2022chain}. Second, the \texttt{VLRMs} category consists of R1-Onevision~\citep{yang2025r1}, Vision-R1~\citep{huang2025vision}, ThinkLite-VL~\citep{wang2025sota}, OpenVLThinker~\citep{deng2025openvlthinker}, and VisionMasters~\citep{li2025vision}. These methods primarily focus on enhancing reasoning capabilities through reinforcement learning or specific fine-tuning. While they possess strong logical deduction skills, they typically lack explicit tool-use mechanisms to acquire fine-grained information through tool interaction. Third, the \texttt{MRAGs} category includes VisDom~\citep{suri2024visdom}, MM-Search-R1~\citep{wu2025mmsearch}, and EVisRAG~\citep{sun2025visrag}. These models are explicitly optimized with multimodal retrieval-augmented generation objectives, aiming to effectively process and integrate information from retrieved document contexts.
Finally, the \texttt{Tool-Enhanced VLMs} are evaluated in our experiments and serve as our primary baselines: Pixel-Reasoner~\citep{wang2025pixel} and VRAG-RL~\citep{wang2025vrag}. These methods incorporate explicit tool execution or action-based mechanisms to support complex visual reasoning, representing the current mainstream approach of using external aids for visual tasks.

\textbf{Implementation Details.}
We employ Qwen2.5-VL-7B-Instruct~\citep{bai2025qwen2} as the backbone model and conduct reinforcement learning using the EasyR1 framework~\citep{zheng2025easyr1}. Training is performed with the DAPO~\cite{yu2025dapo} algorithm, utilizing a group size of 8. For retrieval, we use ColQwen2~\citep{faysse2024colpali} to extract the top-3 document pages per query as visual evidence. Regarding the training data, we collect approximately 11K samples from the training splits of OpenDocVQA~\citep{tanaka2025vdocrag} and SlideVQA~\citep{tanaka2023slidevqa}, applying a rigorous filtering strategy to exclude low-quality or overly simple instances. Additional details on hyperparameters, training environments, and data filtering criteria are provided in Appendices~\ref{app:data-filtering} and \ref{app:implementation-details}.

%% file: section/5_result.tex
\section{Evaluation Results}
\input{table/overall}
This section first presents a comprehensive evaluation of \method{} on three challenging document-based VQA benchmarks. We then conduct ablation studies to validate the effectiveness of individual components of \method{} and provide an in-depth analysis of the role of linguistic tools in enhancing visual perception capabilities.

\subsection{Overall Performance}
As shown in Table~\ref{tab:overall}, we compare \method{} against several baseline models, including prompting-based models, VLRMs, MRAG systems, and Tool-enhanced VLMs. 

Overall, \method{} consistently outperforms all baseline models by achieving improvements of over 4\% in different testing scenarios, which demonstrates its effectiveness in addressing a wide range of visual QA tasks. Compared with prompting-based models, \method{} achieves more than 10\% gains, indicating that our training method enables the backbone model to perform more effective reasoning trajectories for visual understanding and answer questions more accurately.
Based on reinforcement learning, VLRMs such as ThinkLite-VL~\citep{wang2025sota} and OpenVLThinker~\citep{deng2025openvlthinker} also learn to conduct more effective reasoning for answering given questions. Even using the same reinforcement learning strategy, \method{} shows substantial improvements, demonstrating that leveraging carefully designed linguistic tools allows the VLM to converge to more effective reasoning trajectories after training.
While MRAG models shift the focus from enhancing visual reasoning capabilities to evidence denoising and extraction, \method{} outperforms them, highlighting its effectiveness in query-related evidence selection and extraction. Finally, compared with Tool-enhanced baseline models that utilize image-based tools for visual perception, \method{} achieves over 5\% improvements. This demonstrates the advantage of our self-emergent linguistic toolchain, which enables flexible, context-aware, and fine-grained visual operations, rather than relying on rigid external APIs, thereby better facilitating visual perception.

\input{table/ablation}
\input{table/zeroshot}
\input{figure/analyse_attention}
\subsection{Ablation Study}
These experiments comprehensively validate the effectiveness of our two training strategies, namely action RL and tool-based RL, which respectively encourage the VLM to perform more visual understanding actions and to fully exploit the provided linguistic tools. To further evaluate the generalization capability of \method{}, we conduct experiments on both Qwen2.5-VL-3B and 7B backbones~\citep{bai2025qwen2}.

As shown in Table~\ref{tab:ablation-2stage-full}, compared with vanilla LLMs, \method{} yields consistent improvements of over 8\% across different parameter scales (3B and 7B), demonstrating both its effectiveness and strong generalization capability. These consistent gains further indicate that our linguistic toolchain provides a generalizable solution for enhancing the visual perception capability of VLMs, largely independent of model capacity. When removing either the action RL or the tool-based RL phase, the performance of \method{} drops by more than 1\%, indicating that both training strategies play a critical role in enabling fine-grained visual reasoning. While Vanilla DAPO improves the performance of vanilla LLMs by around 5\% through enhancing the standard \texttt{think-then-answer} chain-of-thought reasoning paradigm via RL~\citep{yu2025dapo}, it is still outperformed by our \method{} w/o Tool-based RL variant. This comparison highlights that the self-driven action exploration in the first stage can spontaneously discover visual grounding patterns that are more effective than generic reasoning thoughts. Furthermore, although directly optimizing LLMs to ground the linguistic tools (\method{} w/o Action RL) yields competitive results, it still underperforms the full \method{} framework. This remaining gap confirms that the initial exploration stage not only provides essential reasoning priors for curating higher-quality linguistic tools but also acts as an effective initialization that maximizes the effectiveness of subsequent linguistic tool-enhanced optimization.

\input{figure/case}
\subsection{Zero-Shot Prompt Comparison Across Backbone Scales}
To clarify whether the gains of \method{} already emerge from the prompt format itself, we compare Vanilla prompting, Structured CoT prompting, and the \method{} prompt under a strictly controlled zero-shot setting. Specifically, all results in this comparison are obtained on the same frozen backbones without any RL or fine-tuning. This setting allows us to isolate the contribution of the toolized prompting format from that of subsequent training.

As shown in Table~\ref{tab:zeroshot_prompt_scale}, \method{} consistently outperforms both Vanilla prompting and Structured CoT across all three backbone scales, namely Qwen2.5-VL-3B, Qwen2.5-VL-7B, and Qwen2.5-VL-32B~\citep{bai2025qwen2}. For example, on the 7B backbone, \method{} achieves an average score of 56.19, compared with 54.43 for Structured CoT and 53.12 for Vanilla prompting. Similar gains are also observed on the 3B and 32B backbones. These results indicate that the toolized \method{} format itself provides additional value beyond a standard natural-language step-by-step prompt, even before training is applied.

\subsection{Effectiveness of Linguistic Tools in Visual Document Perception}
To investigate the underlying mechanisms behind the performance gains of \method{}, we conduct a comprehensive quantitative analysis, as shown in Figure~\ref{fig:attention_comparison}, to examine how linguistic tools enhance the visual perception capability of backbone models through curated linguistic supervision.

First, we employ a vanilla LLM to analyze the relationship between perception rate and QA accuracy. As illustrated in Figure~\ref{fig:attn_vanilla}, the average accuracy of the vanilla model exhibits a strong positive correlation with the attention hit rate. This empirical observation highlights the critical role of an accurate visual perception in capturing key information from the given document pages. Motivated by this finding, we further plot the perception rate together with QA accuracy of different models in Figure~\ref{fig:attn_vrag}. The results consistently show that QA accuracy improves as the perception rate increases. Notably, \method{} demonstrates its effectiveness by achieving the best performance in terms of both QA accuracy and perception rate. In contrast, the tool-enhanced VLMs, Pixel-Reasoner and VRAG-RL, exhibit nearly identical perception rates while attaining different QA accuracies. This suggests that image-tool-based methods may struggle to precisely capture key visual regions when relying on raw image operations, such as clipping.

We then conduct deeper analyses to further validate the effectiveness of different models. As shown in Figure~\ref{fig:V-precision}, we report the QA accuracy on queries where models successfully attend to the golden regions of visual documents. Higher accuracy indicates a stronger ability to exploit visual evidence for question answering. \method{} achieves the highest score, demonstrating its superiority in fully leveraging visual clues in the given image to generate more accurate answers. Furthermore, Figure~\ref{fig:V-recall} presents the perception rate for queries that are answered correctly. A lower rate suggests that the answer generation relies more heavily on memorized knowledge, which may increase the risk of hallucination. \method{} again achieves the highest perception rate, indicating its potential to alleviate knowledge conflicts and encourage stronger reliance on external visual information.

\subsection{Case Study}
To empirically demonstrate the effectiveness of \method{}, we randomly select one representative case from SlideVQA for qualitative analysis. We compare \method{} with a vanilla VLM and VRAG-RL. VRAG-RL relies on explicit image tools for image processing, whereas \method{} leverages linguistic tools to guide visual attention toward relevant evidence, enabling more fine-grained perception.

As illustrated in Figure~\ref{fig:case-study}, the user queries the specific percentage of people who never carry cash. The vanilla VLM model attends to the ground-truth region but distributes its attention across multiple irrelevant regions in an attempt to gather additional visual evidence, which ultimately misleads the model and prevents it from accurately answering the question. In contrast, \method{} concentrates the VLM's attention on the ground-truth region and accurately generates the golden answer, demonstrating its effectiveness in enhancing the perceptual capability of VLMs. On the other hand, VRAG-RL~\citep{wang2025vrag} attempts to resolve the ambiguity of visual understanding through active cropping and answers the question based on salient evidence retained after image tool execution. However, in this case, VRAG-RL performs an incorrect action by cropping out the crucial information ``43\%'', leading to an erroneous answer of ``56\%''. This example illustrates that enhancing visual perception through explicit image tools may introduce the risk of incorrect image operations, resulting in the loss of critical visual information.

%% file: table/overall.tex
\begin{table*}[t]
\centering

\resizebox{1.0\textwidth}{!}{\begin{tabular}{lccccccccccccc}
\toprule
\multirow{3}{*}{\textbf{Methods}}
& \multicolumn{3}{c}{\textbf{In Domain}} 
& \multicolumn{9}{c}{\textbf{Out of Domain}} 
& \multirow{3}{*}{\textbf{Avg.}} \\
\cmidrule(lr){2-4}\cmidrule(lr){5-13}
& \multicolumn{3}{c}{\textbf{SlideVQA}} 
& \multicolumn{6}{c}{\textbf{MMLongBench-Doc}} 
& \multicolumn{3}{c}{\textbf{ViDoSeek}} 
& \\
\cmidrule(lr){2-4}\cmidrule(lr){5-10}\cmidrule(lr){11-13} 
& Single & Multi & Overall 
& TXT & TAB & CHA & FIG & LAY & Overall 
& Single & Multi & Overall 
& \\
\midrule

% ===== Base 1 =====
\rowcolor{gray!20}\multicolumn{14}{l}{\textbf{\textit{Prompting Methods}}} \\\midrule
Direct
& 71.36 & 52.56 & 66.55
& 31.21 & 23.50 & 27.53 & 23.59 & 21.19 & 27.59
& 64.50 & 66.20 & 65.24
& 53.12 \\
TOT
& 66.02 & 39.68  & 59.28 
& 27.18 & 22.12 & 21.91 & 27.57 & 25.42 & 27.82
& 57.36 & 69.82 & 62.78 
& 49.96  \\
GOT
& 69.11 & 45.68  & 63.12 
& 29.19 &21.66& 22.47 & 24.25 & 22.03 & 27.12
& 55.66 & 62.37 & 58.58 
&  49.60 \\\midrule
\rowcolor{gray!20}\multicolumn{14}{l}{\textbf{\textit{Vision-Language Reasoning Models (VLRMs)}}} \\\midrule
R1-Onevision
& 71.78 &51.32   & 66.55
& 30.54 &26.73  &27.53  &27.57 & 25.42 &30.50 
& 62.48 &68.41   &65.06  
& 54.03  \\
Vision-R1 
& 78.09 &53.79   &71.87  
&32.55&27.65  &26.40  &29.90 & \underline{30.51} &32.25 
& 60.31 &68.01  &63.66  
& 55.92  \\
ThinkLite-VL
& 76.88 &\underline{58.02}   &72.05  
& 33.22 &24.88  &26.97  &29.90 & 27.12 &31.66 
& 63.88 &71.43  &67.16  
& 56.96  \\
OpenVLThinker
& 78.94 &57.50   &\underline{73.45}  
& 35.23 &\underline{28.57}  &26.97  &30.23 & 27.97 &32.71 
& 65.43 &\underline{74.45}  &69.35  
& \underline{58.50}  \\
VisionMatters
& 74.76 &56.08   &69.98  
& 35.91 &23.96  &29.21  &27.91 & 28.81 & 31.66
& 61.40 &72.23  &66.11  
&  55.91 \\\midrule

\rowcolor{gray!20}\multicolumn{14}{l}{\textbf{\textit{Multimodal Retrieval-Augmented Generation Models (MRAGs)}}} \\\midrule
VisDom
& 73.79 & 50.09 & 67.72
& 33.89 & 21.20 & 21.35 & 18.94 & 21.19 & 25.96
& 62.95 & 70.62 & 66.29
& 53.32 \\
MM-Search-R1
& 76.64 &53.44  &70.70 
& 31.54 &25.35  &27.53  &26.91  &22.88  &29.92 
& 65.58 &72.23  &68.48 
& 56.36\\
EVisRAG
& 79.55 &54.67   &73.18  
& 32.55 &27.19  &25.28  &27.57 & 27.12 &30.97 
& 66.51 & 73.84 & \underline{69.70}  
& 57.95  \\\midrule
\rowcolor{gray!20}\multicolumn{14}{l}{\textbf{\textit{Tool-Enhanced VLMs}}} \\\midrule
Pixel-Reasoner
& 78.13 &57.50  &72.84
&\underline{37.37}  & 27.19 &24.72  &\textbf{32.89} &27.97  &\underline{33.22} 
& \underline{67.96} & 70.1 &68.89
& 58.31 \\
VRAG-RL
& \underline{81.74} & 49.21 & 73.41
& 31.54 & 24.88 &\underline{30.34} & 30.23 & 20.34 & 31.55
& 64.65 &71.63 &67.69 
& 57.55 \\
\textbf{\method{}}
& \textbf{83.62} &\textbf{64.55}   &\textbf{78.74}  
& \textbf{38.59} & \textbf{31.80} &\textbf{31.46}  &\underline{32.23} & \textbf{33.05} &\textbf{36.55}
& \textbf{74.25} &\textbf{75.35}  & \textbf{74.87}
& \textbf{63.39} \\

\bottomrule
\end{tabular}}
\caption{
Overall performance. The best results are highlighted in \textbf{bold}, and the second-best are \underline{underlined}.
}
\label{tab:overall}
\end{table*}

%% file: table/ablation.tex
\begin{table}[t]
\centering

\resizebox{\columnwidth}{!}{%
\begin{tabular}{lcccc}
\toprule
\textbf{Methods} & \textbf{SlideVQA} & \textbf{ViDoSeek} & \textbf{MMBench} & \textbf{Avg.} \\
\midrule

\multicolumn{5}{l}{\textit{\textbf{Qwen2.5-VL-3B-Instruct}}} \\ 
\midrule
Vanilla VLM & 60.18 & 63.80 & 25.61 & 49.86 \\
Vanilla DAPO & 67.72 & 67.42 & 29.10 & 54.74 \\
\method{} & \textbf{71.02} & \textbf{71.98} & \textbf{31.20} & \textbf{58.06} \\
w/o Action RL & 69.03 & 69.18 &30.27 & 56.22 \\
w/o Tool-based RL & 68.35 & 68.30 & 29.68 & 55.47 \\
\midrule

\multicolumn{5}{l}{\textit{\textbf{Qwen2.5-VL-7B-Instruct}}} \\ 
\midrule
Vanilla VLM & 66.55 & 65.24 & 27.59 & 53.12 \\
Vanilla DAPO & 76.79 & 70.23 & 35.86 & 60.96 \\
\method{} & \textbf{78.74} & \textbf{74.87} & \textbf{36.55} & \textbf{63.39} \\
w/o Action RL & 77.25 & 73.23 & 35.39 & 61.95 \\
w/o Tool-based RL & 76.03 & 72.68 & 35.86 & 61.52 \\
\midrule
\end{tabular}%
}
\caption{Ablation results comparing different training strategies used by \method{}.}

\label{tab:ablation-2stage-full}
\end{table}

%% file: table/zeroshot.tex
\begin{table}[t]
\centering

\resizebox{\columnwidth}{!}{%
\begin{tabular}{lcccc}
\toprule
\textbf{Methods} & \textbf{SlideVQA} & \textbf{ViDoSeek} & \textbf{MMBench} & \textbf{Avg.} \\
\midrule

\multicolumn{5}{l}{\textit{\textbf{Qwen2.5-VL-3B-Instruct}}} \\
\midrule
Vanilla VLM & 60.18 & 63.80 & 25.61 & 49.86 \\
Structured CoT & 61.42 & 65.07 & 26.54 & 51.01 \\
\method{} & \textbf{62.60} & \textbf{66.40} & \textbf{27.80} & \textbf{52.27} \\
\midrule

\multicolumn{5}{l}{\textit{\textbf{Qwen2.5-VL-7B-Instruct}}} \\
\midrule
Vanilla VLM & 66.55 & 65.24 & 27.59 & 53.12 \\
Structured CoT & 67.76 & 66.98 & 28.55 & 54.43 \\
\method{} & \textbf{68.67} & \textbf{69.61} & \textbf{30.30} & \textbf{56.19} \\
\midrule

\multicolumn{5}{l}{\textit{\textbf{Qwen2.5-VL-32B-Instruct}}} \\
\midrule
Vanilla VLM & 78.74 & 76.44 & 39.35 & 64.84 \\
Structured CoT & 79.30 & 77.20 & 39.90 & 65.47 \\
\method{} & \textbf{80.10} & \textbf{78.40} & \textbf{40.80} & \textbf{66.43} \\
\midrule

\end{tabular}%
}
\caption{Zero-shot prompt comparison across different backbone scales. All methods are evaluated on frozen backbones without RL or fine-tuning. \method{} consistently outperforms Vanilla prompting and Structured CoT across 3B, 7B, and 32B models.}
\label{tab:zeroshot_prompt_scale}
\end{table}

%% file: figure/analyse_attention.tex
\begin{figure}[t]
    \centering
    \begin{subfigure}[t]{0.48\linewidth}
        \centering
        \includegraphics[width=\linewidth]{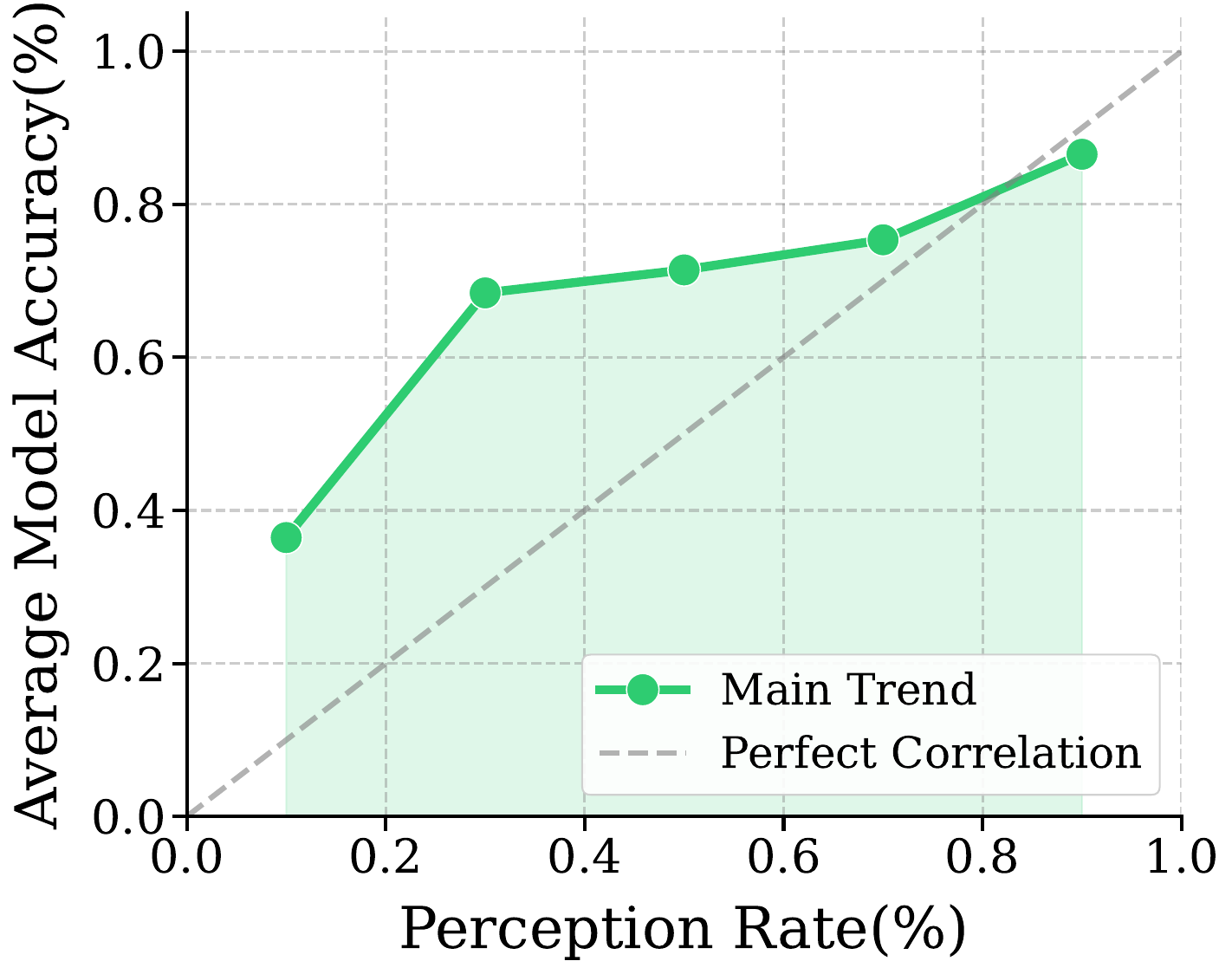}
        \caption{Correlation between the golden-region perception rate and QA accuracy.}
        \label{fig:attn_vanilla}
    \end{subfigure}
    \hfill
    \begin{subfigure}[t]{0.48\linewidth}
        \centering
        \includegraphics[width=\linewidth]{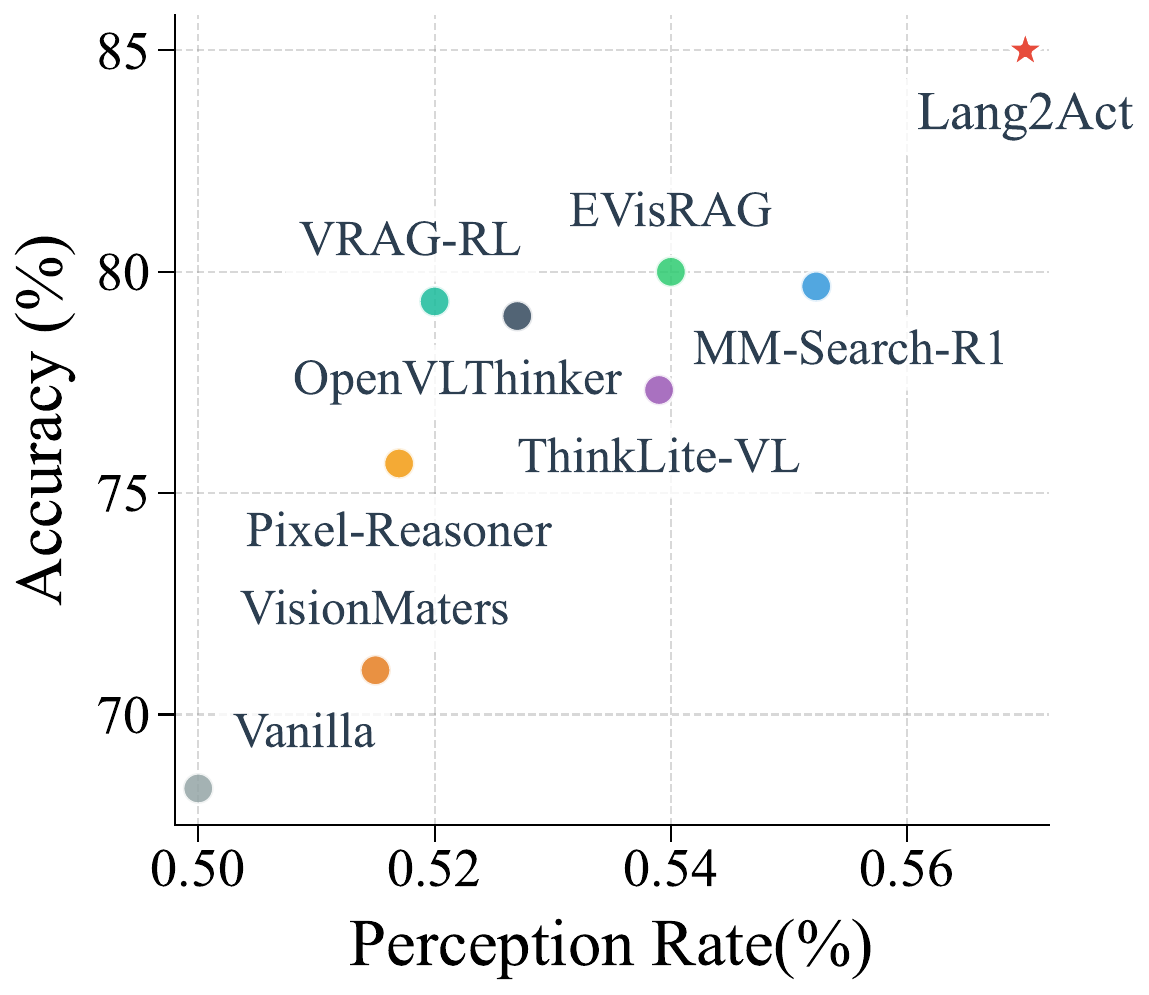}
        \caption{QA accuracy and relative perception rate across different methods.}
        \label{fig:attn_vrag}
    \end{subfigure}

    \vspace{0.5em}

    \begin{subfigure}[t]{0.48\linewidth}
        \centering
        \includegraphics[width=\linewidth]{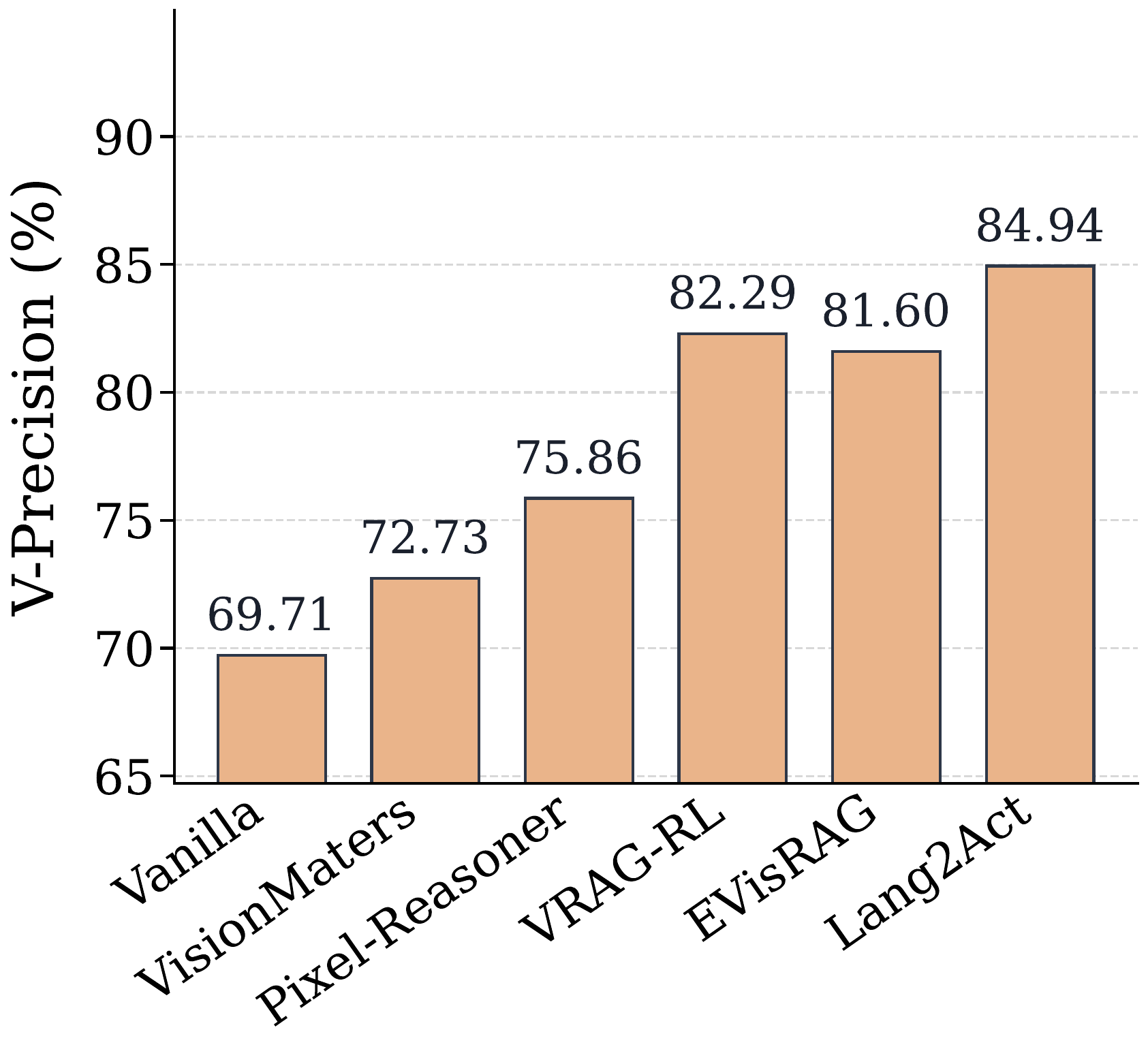}
        \caption{QA accuracy when models successfully perceive the golden regions.}
        \label{fig:V-precision}
    \end{subfigure}
    \hfill
    \begin{subfigure}[t]{0.48\linewidth}
        \centering
        \includegraphics[width=\linewidth]{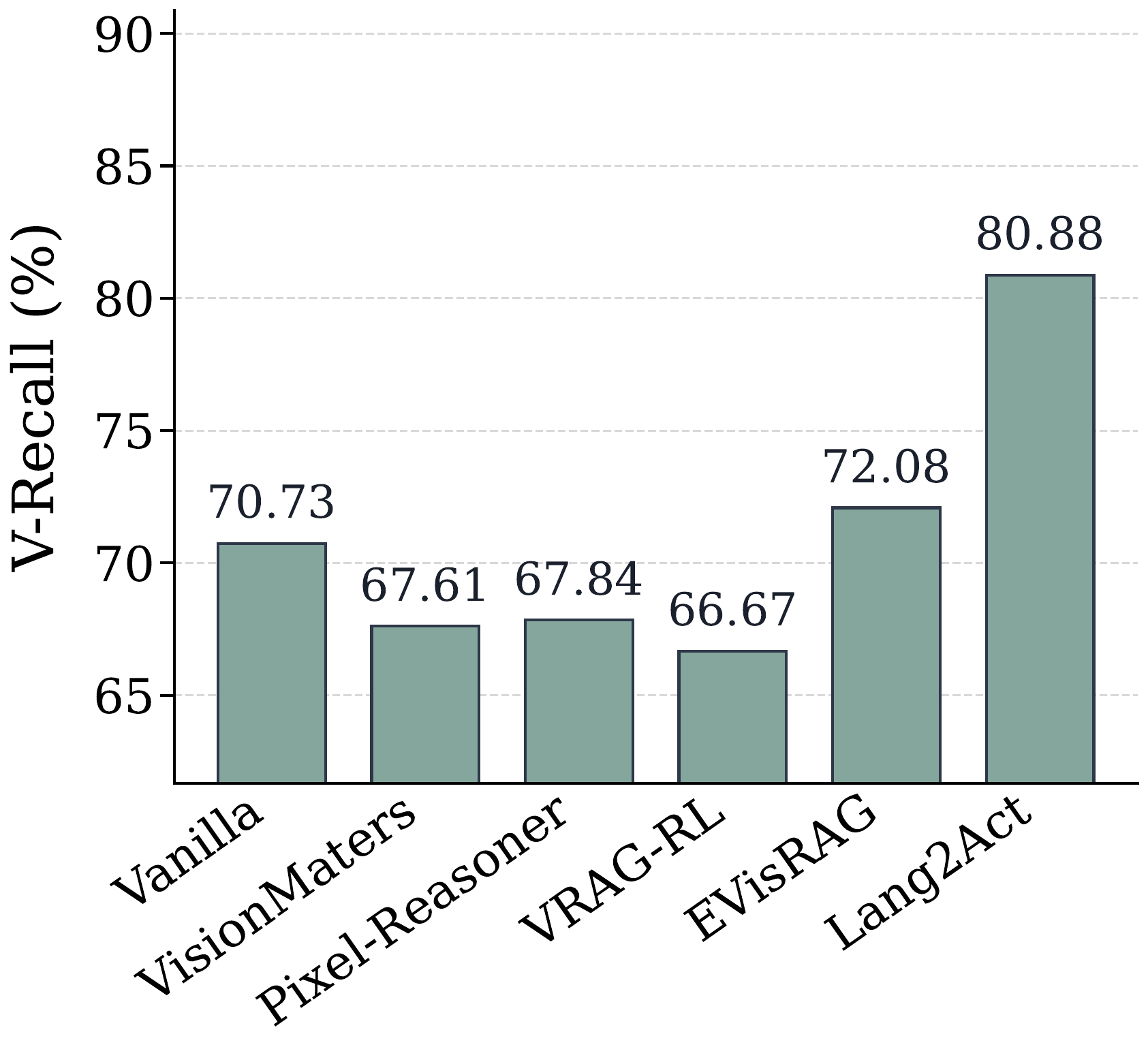}
        \caption{Golden-region perception ratio for correctly answered questions.}
        \label{fig:V-recall}
    \end{subfigure}

    \caption{Quantitative analysis of image perception quality in relation to QA accuracy. We compute V-Precision (Figure~\ref{fig:V-precision}) and V-Recall (Figure~\ref{fig:V-recall}) for analysis by leveraging the perception rate on the golden region together with QA accuracy. The perception rate is calculated according to whether the model's internal attention hit the golden region.}
    \label{fig:attention_comparison}
\end{figure}

%% file: figure/case.tex
\begin{figure*}[t]
    \centering
    \includegraphics[width=\textwidth]{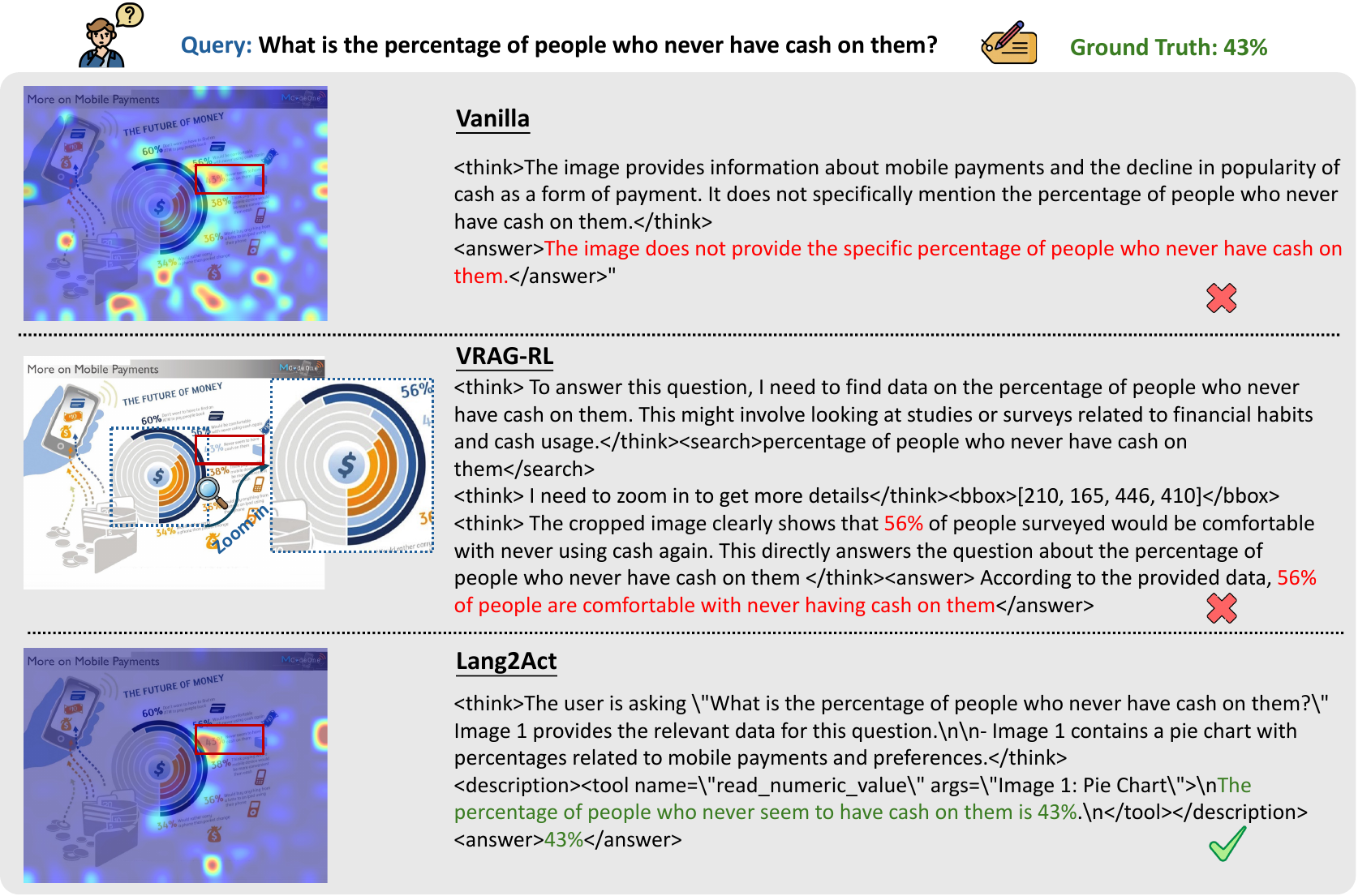}
    \caption{
    Case Study on SlideVQA. The red box indicates the ground truth region of the given image.}
    \label{fig:case-study}
\end{figure*}

%% file: section/6_conclusion.tex
\section{Conclusion}
This paper proposes \method{} to leverage self-emergent linguistic toolchains for fine-grained visual perception. Experimental results demonstrate the effectiveness of \method{}, which further internalizes visual actions to bridge the gap between reasoning and fine-grained visual perception.

\section*{Limitations}

\method{} demonstrates superior effectiveness and efficiency compared to existing tool-enhanced VLMs, particularly in enabling fine-grained visual perception with an intrinsic linguistic toolbox. Our approach successfully concentrates the model's visual attention onto informative regions through intrinsic linguistic tools, a behavior that empirically aligns with improved answer accuracy. However, fully disentangling the strict causal dynamics between these attentional shifts and the final generation outcomes remains a complex challenge, primarily due to the inherent black-box nature of neural networks. While our current experimental analysis establishes a robust positive correlation between attention concentration and reasoning correctness, the theoretical formalization of this causality presents an open avenue for further exploration. 

%% file: section/7_acknowledge.tex
\section*{Acknowledgments}
This work was supported by Alibaba Group through Alibaba Innovative Research Program.
% This document has been adapted
% by Steven Bethard, Ryan Cotterell and Rui Yan
% from the instructions for earlier ACL and NAACL proceedings, including those for
% ACL 2019 by Douwe Kiela and Ivan Vuli\'{c},
% NAACL 2019 by Stephanie Lukin and Alla Roskovskaya,
% ACL 2018 by Shay Cohen, Kevin Gimpel, and Wei Lu,
% NAACL 2018 by Margaret Mitchell and Stephanie Lukin,
% Bib\TeX{} suggestions for (NA)ACL 2017/2018 from Jason Eisner,
% ACL 2017 by Dan Gildea and Min-Yen Kan,
% NAACL 2017 by Margaret Mitchell,
% ACL 2012 by Maggie Li and Michael White,
% ACL 2010 by Jing-Shin Chang and Philipp Koehn,
% ACL 2008 by Johanna D. Moore, Simone Teufel, James Allan, and Sadaoki Furui,
% ACL 2005 by Hwee Tou Ng and Kemal Oflazer,
% ACL 2002 by Eugene Charniak and Dekang Lin,
% and earlier ACL and EACL formats written by several people, including
% John Chen, Henry S. Thompson and Donald Walker.
% Additional elements were taken from the formatting instructions of the \emph{International Joint Conference on Artificial Intelligence} and the \emph{Conference on Computer Vision and Pattern Recognition}.

%% file: section/8_appendix.tex
\appendix
\section{Appendix}
\subsection{License}
We summarize the licenses and usage terms of the datasets used in this work. 
ViDoseek and MMLongBench-Doc are released under the Apache 2.0 license. 
SlideVQA and OpenDocVQA are provided under the NTT Software Evaluation License, which permits non-commercial academic use for research and evaluation purposes. 
We strictly follow the original licensing terms of all datasets and do not redistribute any third-party raw data.

\subsection{Additional Details of Datasets}
\label{app:datasets}
To comprehensively evaluate \method{}'s ability to maintain a continuous reasoning context and mitigate visual hallucinations, we conduct experiments on three representative benchmarks covering diverse document scenarios. We first employ SlideVQA~\citep{tanaka2023slidevqa} to assess cross-slide reasoning over interconnected text and diagrams, serving as a rigorous testbed for aggregating fragmented visual evidence without the context loss often induced by rigid cropping. To further validate fine-grained perception in dense layouts, we utilize ViDoSeek~\citep{wang2025vidorag}, which challenges the model to precisely capture specific attributes in visually rich documents where raw image operations often fail. Additionally, we incorporate MMLongBench-Doc~\citep{ma2024mmlongbench} to examine performance on long multimodal documents, ensuring our method sustains accurate visual grounding over extended horizons and effectively alleviates the risk of hallucination caused by error accumulation.
Table~\ref{tab:dataset_stats} summarizes the query counts and dataset characteristics.
\subsection{Experimental Details of Data Filtering}
\label{app:data-filtering}
In the Tool-Based Optimization training, we employ the model obtained from the  Action Exploration as the initialization. 
For each training sample, the model generates eight candidate completions through stochastic sampling. 
Samples for which all eight completions are correct are removed, as they provide a limited learning signal for further optimization. 
After filtering, approximately 5,700 samples are retained to form the training set for the second-stage RL, focusing the optimization on more challenging instances. 
Figure~\ref{fig:difficulty} illustrates the distribution of sample difficulty before and after filtering.

\subsection{Additional Implementation Details}
\label{app:implementation-details}
All experiments are conducted on NVIDIA A800 GPUs. The detailed hyperparameters we use during the training period of Action RL and Tool-based RL are shown in Table~\ref{tab:grpo-hparams} and Table~\ref{tab:grpo-hparams2}.

\input{table/datasets}
\input{figure/data_filter} 
\textbf{Reward Function for DAPO Training.}
We follow the reward formulation defined in Eq.~\ref{eq:reward}.
In all experiments, we set $\alpha = 0.8$ and $\beta = 0.2$.
The answer reward $r_{\text{ans}}(z, a)$ evaluates whether the predicted answer is correct.
Following Eq.~\ref{eq:dapo_1}, we adopt an automatic evaluator to compare the
generated answer $a$ , which is extracted from the \texttt{<answer>} block, with the ground-truth answer $a^*$, and assign a binary score:
\begin{equation}\small
r_{\text{ans}}(z,a) =
\begin{cases}
1, & \text{if the generated answer is correct }. \\
0, & \text{otherwise.}
\end{cases}
\end{equation}

Beyond the answer-based reward, we also incorporate a tool reward that enforces structured reasoning by requiring the model to generate outputs following a fixed tag order, namely \texttt{<think>}, \texttt{<description>}, and \texttt{<answer>}.
In particular, linguistic tool invocations are constrained to appear only
within the \texttt{<description>} block and must conform to the curated toolbox,
thereby encouraging disciplined and well-structured tool usage during visual
reasoning.

\textbf{Retrieval Implementation Details.}
We use ColPali~\citep{faysse2024colpali} as the visual embedding model to encode document pages and queries for retrieval. 
We build and query the retrieval index with LlamaIndex using similarity search. 
Unless otherwise specified, we retrieve the top-3 pages for each query and provide the same retrieved evidence to all methods for evaluation. 
Table~\ref{tab:colqwen2_retrieval_all} reports the retrieval performance of our retriever across the evaluated benchmarks.

\input{table/stage1-hyperparameters}
\input{table/stage2-hyperparameters}
\input{table/retrieval}
\textbf{Baselines and Comparison Setup.}
We compare our method with a diverse set of strong baselines covering vision-language reasoning models and retrieval-augmented generation approaches.

R1-Onevision-7B~\citep{yang2025r1} proposes a unified multimodal reasoning framework by formally aligning visual and textual representations. It leverages reinforcement learning to improve cross-modal reasoning without relying on task-specific heuristics.

Vision-R1-7B~\citep{huang2025vision} further extends reinforcement learning to multimodal large language models by introducing vision-guided reward signals. This approach incentivizes step-by-step reasoning grounded in visual perception, eliminating the need for human-annotated preference data.

OpenVLThinker-7B~\citep{deng2025openvlthinker} investigates iterative self-improvement for LVLM reasoning by alternating SFT and GRPO-style RL, showing that SFT can surface useful reasoning behaviors and narrow the search space for subsequent RL, leading to stronger multi-step visual reasoning.

ThinkLite-VL~\citep{wang2025sota} adopts a lightweight reasoning-oriented training strategy that emphasizes efficient self-improvement. It alternates between supervised fine-tuning and reinforcement learning, enabling the model to progressively refine its multimodal reasoning capability with reduced computational overhead.

VisionMatters~\citep{li2025vision} revisits multimodal reasoning from the perspective of image perturbation. Systematically analyzing how visual variations affect model predictions, it enhances robustness and visual sensitivity through targeted fine-tuning.

Pixel-Reasoner~\citep{wang2025pixel} explicitly encourages fine-grained pixel-space reasoning via curiosity-driven reinforcement learning.

MM-Search-R1~\citep{wu2025mmsearch} equips multimodal models with explicit search capabilities, enabling iterative retrieval and reasoning over external visual evidence.

EVisRAG~\citep{sun2025visrag} addresses the challenge of multi-image integration in VRAG systems. It proposes an evidence-guided paradigm trained via Reward-Scoped GRPO (RS-GRPO), which incentivizes the model to explicitly extract evidence from individual images before synthesizing the final answer

VRAG-RL~\citep{wang2025vrag} further integrates reinforcement learning into the RAG paradigm, optimizing vision-perception-driven retrieval and reasoning through iterative policy improvement.

\input{table/acc}
\subsection{Additional Baseline Comparison Results}
In addition to the LLM-based evaluation reported in the main paper, we provide supplementary results using an automatic accuracy metric in Table~\ref{tab:main_overall_acc}. 
This evaluation directly compares predicted answers with ground-truth responses to compute exact-match accuracy, without relying on an external judge model.

The overall performance trends of accuracy remain consistent with those evaluated in LLM-as-judge in Table~\ref{tab:overall}, demonstrating that the improvements of \method{} are not sensitive to the choice of evaluation protocol. 

\subsection{Tool Frequency of the Curated Toolbox}
To analyze the behavioral patterns emerging from the self-driven exploration phase, we sampled 1,500 reasoning trajectories and aggregated the usage frequency of each linguistic tool. The statistical distribution is presented in Table~\ref{tab:tool_frequency}. The results reveal a clear hierarchy in visual reasoning. The most frequently employed tools are Perception-Oriented actions, specifically \texttt{read\_text\_element} (64.26\%) and \texttt{read\_numeric\_value} (41.73\%). This dominance indicates that the model prioritizes precise information extraction as the foundation for answering document-based queries. Following perception, reasoning-oriented tools such as \texttt{identify\_entity\_attribute}, \texttt{compare\_values}, and \texttt{locate\_visual\_element} exhibit stable usage frequencies, demonstrating the model's capability to perform structural analysis and comparative reasoning after grounding the visual evidence. Based on this frequency distribution, we observe a significant long-tail effect. Tools ranked 8th and below (e.g., specific arithmetic operations like \texttt{subtract\_values}) appear with negligible frequency, representing outlier cases that contribute little to generalizability. Consequently, to construct a compact and efficient linguistic toolbox $\mathcal{T}_{box}$, we selected the top-7 most frequent tools. This selection covers over 98.9\% of the total tool usage observed in the sampled trajectories, ensuring that the final toolbox encapsulates the core visual reasoning primitives while filtering out sparse, task-specific noise.

\input{table/Tool_Statistics}
\input{figure/cot_analyse}

\subsection{Reliability of Tool Curation and Tool-Integrated Trajectory Generation}
To further clarify the reliability of the tool curation process and the model's ability to generate valid tool-integrated trajectories, we provide an additional empirical analysis on tool-related errors before and after training. In our pipeline, the tool curation procedure is fully prompt-driven: we first prompt the model to abstract generated reasoning trajectories into structured, atomic cognitive operations, and then prompt it to extract linguistic tools from these abstractions. During extraction, the model is provided with the current tool pool and instructed to reuse an existing tool whenever the operation matches its functional semantics and argument schema; otherwise, it defines a new tool following a unified format. This reuse-before-creation strategy helps maintain consistency and prevents uncontrolled expansion of the tool space. The exact system prompt used for tool curation is shown in Figure~\ref{fig:tools_curation_prompt}.

To examine whether Qwen2.5-VL can reliably produce tool-included trajectories, we analyze tool-related errors on 2,215 trajectories before and after training. We categorize the errors into three types: \textit{Format Error}, where the generated tool invocation does not follow the required structured format; \textit{Tool Not Contain}, where the generated tool is not included in the curated toolbox; and \textit{No Tool}, where the model fails to produce any tool invocation when one is expected.

Table~\ref{tab:tool_reliability} reports the results. The vanilla Qwen2.5-VL-7B model exhibits a 9.35\% overall tool-related error rate when generating trajectories. After training, Lang2Act reduces the overall error rate to 0.27\%, with both \textit{Format Error} and \textit{Tool Not Contain} reduced to zero. These results indicate that our training significantly improves the model's ability to produce valid tool-augmented trajectories and consistently follow the required tool format, thereby making the approach more reliable and easier to reproduce in practice.

\begin{table}[t]
\centering
\small
\setlength{\tabcolsep}{6pt}
\begin{tabular}{lcc}
\toprule
\textbf{Error Type} & \makecell{\textbf{Vanilla} \\ \textbf{(Count/\%)}} & \makecell{\textbf{Lang2Act} \\ \textbf{(Count/\%)}} \\
\midrule
Format Error     & 87 (3.93\%)  & 0 (0\%) \\
Tool Not Contain & 24 (1.07\%)  & 0 (0\%) \\
No Tool          & 96 (4.35\%)  & 6 (0.27\%) \\
\midrule
Total Errors     & 207 (9.35\%) & 6 (0.27\%) \\
\bottomrule
\end{tabular}
\caption{Tool-related error analysis on 2,215 trajectories before and after training. Lang2Act substantially reduces tool generation errors and improves the reliability of producing valid tool-augmented reasoning trajectories.}
\label{tab:tool_reliability}
\end{table}

\subsection{Toolbox Generalization Across Datasets}
To further evaluate whether the curated toolbox is overly tied to the source dataset or prompt distribution, we additionally assess \method{} on two out-of-domain table-document VQA benchmarks, namely FetaTab and PaperTab. We directly reuse the same toolbox without any re-ranking, re-selection, or modification. This setting provides a direct test of whether the linguistic tools discovered from the source distribution can transfer to unseen document domains.

Table~\ref{tab:toolbox_generalization} reports the results. \method{} achieves the best performance on both benchmarks, obtaining 55.31 on FetaTab and 24.94 on PaperTab, outperforming all compared baselines. These results provide additional evidence that the curated toolbox is transferable across datasets and prompt distributions, rather than being narrowly specialized to the source training distribution.

\begin{table}[t]
\centering
\small
\setlength{\tabcolsep}{8pt}
\begin{tabular}{lcc}
\toprule
\textbf{Methods} & \textbf{FetaTab} & \textbf{PaperTab} \\
\midrule
Vanilla         & 51.87 & 19.59 \\
R1-Onevision    & 49.80 & 16.79 \\
Vision-R1       & 47.64 & 20.87 \\
OpenVLThinker   & 49.31 & 20.61 \\
VisionMatters   & 53.25 & 21.63 \\
MM-Search-R1    & 50.49 & 20.36 \\
EVisRAG         & 53.54 & 22.90 \\
Pixel-Reasoner  & 51.77 & 22.17 \\
VRAG-RL         & 44.49 & 20.10 \\
\textbf{\method{}} & \textbf{55.31} & \textbf{24.94} \\
\bottomrule
\end{tabular}
\caption{Cross-dataset generalization results on FetaTab and PaperTab. The same curated toolbox is directly reused without re-ranking, re-selection, or modification.}
\label{tab:toolbox_generalization}
\end{table}

\subsection{Ideal-Condition Comparison with Ground-Truth Bounding Boxes}
To provide a stronger upper-bound comparison for image-tool-based methods, we further evaluate an ideal-condition variant of VRAG-RL with access to ground-truth bounding boxes. Specifically, we manually annotate 500 examples with gold bounding boxes and allow the image-tool-based variant to directly use these annotations during inference. This setting helps isolate whether linguistic tools still offer advantages even when the region selection error of image-based methods is minimized.

Table~\ref{tab:gt_bbox_comparison} reports the results. Even under this idealized setting, \method{} still substantially outperforms the image-tool-based baselines. In particular, \method{} achieves an accuracy of 86.20, compared with 77.80 for VRAG-RL and 78.00 for Pixel-Reasoner. These results suggest that the advantage of \method{} does not solely come from avoiding imperfect cropping. Instead, the proposed linguistic toolbox more effectively preserves the coupling between fine-grained visual evidence and downstream reasoning, even when image-based methods are given oracle region annotations.

\begin{table}[t]
\centering
\small
\setlength{\tabcolsep}{10pt}
\begin{tabular}{lc}
\toprule
\textbf{Method} & \textbf{Acc} \\
\midrule
Vanilla VLM     & 74.20 \\
VRAG-RL         & 77.80 \\
Pixel-Reasoner  & 78.00 \\
\textbf{\method{}} & \textbf{86.20} \\
\bottomrule
\end{tabular}
\caption{Ideal-condition comparison with ground-truth bounding boxes. We manually annotate 500 examples with gold bounding boxes and provide them to image-tool-based baselines during inference. Even in this oracle-region setting, \method{} still achieves the best performance.}
\label{tab:gt_bbox_comparison}
\end{table}
\subsection{Performance Analysis under Oracle Retrieval}
To rigorously assess the model's fine-grained visual reasoning capabilities and its resilience to interference, we conducted an oracle setting experiment in Table~\ref{tab:performance_comparison}, where ground-truth pages containing the answer are directly provided, supplemented by distractor images to ensure a minimum input of three pages per query.

In the oracle setting, where the correct document is guaranteed, VRAG-RL~\citep{wang2025vrag} exhibits competitive performance, validating that its active cropping mechanism effectively enhances perception by physically zooming into specific regions. However, \method{} consistently surpasses this strong baseline, particularly on detail-intensive benchmarks like ViDoSeek~\citep{wang2025vidorag}. This performance advantage demonstrates that while mechanical cropping improves resolution, it inherently risks severing the semantic link between local details and the global layout, whereas our linguistic toolchain maintains the holistic context required for complex interpretation.
\input{table/analyse1}
% \subsection{Case Studies: \method{} vs. Baselines}
\subsection{Analysis of Model Confidence and Response Quality}
To deeply evaluate the quality of the generated reasoning chains beyond simple accuracy, we employed an advanced LLM judge to assess three critical dimensions: hallucination, factual consistency, and coherence, using the evaluation prompt illustrated in Figure~\ref{fig:llm_judge_reasoning}. As shown in Figure~\ref{fig:quality_comparison}, the results indicate that existing tool-enhanced approaches often struggle with coherence and consistency, primarily due to the context fragmentation caused by rigid raw image operations. By physically isolating visual regions, these methods sever the semantic connection between local details and global structures, frequently forcing the model to hallucinate information to bridge logical gaps. In contrast, \method{} achieves superior performance by leveraging its self-emergent linguistic toolchain to internalize visual perception into the autoregressive generation process. This design maintains a continuous reasoning flow where visual grounding is tightly coupled with logical deduction, effectively suppressing hallucinations and ensuring that the generated trajectories remain factually consistent and logically coherent.
\input{table/latency}
\subsection{Inference Latency of \method{}.}
We evaluate inference latency on the SlideVQA dataset, measuring the end-to-end runtime required to generate a final answer for each query. All methods are evaluated under the same experimental settings to ensure a fair comparison as shown in Table~\ref{tab:latency_comparison}.
\subsection{Prompt Examples}
Below are sample prompts used for multimodal reasoning tasks.
This section provides the specific prompt templates used for the baselines and our method.

Figure~\ref{fig:vanilla2_prompt} presents the Vanilla prompt, which instructs the model to conduct internal reasoning within \texttt{<think>} tags before providing a direct answer, serving as the standard baseline. Figure~\ref{fig:xml_cot_vqa_prompt} illustrates the Action RL prompt, requiring the model to explicitly describe the visual evidence used for reasoning, which is utilized to generate high-quality training data. Figure~\ref{fig:tools_curation_prompt} presents the Tools Curation prompt, which deconstructs these reasoning steps into atomic, structure-aware cognitive operations to construct the tool pool.Figure~\ref{fig:evisrag_prompt} displays the EVisRAG~\citep{sun2025visrag} prompt, which enforces a strict four-step structured reasoning process: observing images, recording evidence, reasoning, and answering. Figure~\ref{fig:lexicon_prompt} depicts the prompt for our proposed Lang2Act framework. It defines a set of linguistic tools (e.g., numerical extraction, visual element identification) in the context and requires the model to perform fine-grained analysis and grounding of visual information using these tools within the \texttt{<description>} tag before generating the final answer.

Regarding complex reasoning strategies, Figure~\ref{fig:tot_prompt} details the Tree-of-Thoughts (ToT) prompt~\citep{yao2023tree}, guiding the model to deconstruct the problem into sub-problems and evaluate the validity of multiple reasoning branches. Figure~\ref{fig:got_prompt} presents the Graph-of-Thoughts (GOT) prompt~\citep{besta2024graph}, asking the model to generate initial thoughts and then refine and merge them to construct a comprehensive reasoning graph. For tool-enhanced methods, Figure~\ref{fig:vrag_rl_prompt} illustrates the VRAG-RL prompt~\citep{wang2025vrag}, which allows the agent to query knowledge via a search engine and execute image cropping using \texttt{<bbox>} tags to acquire local details. Figure~\ref{fig:pixel_reasoner_prompt} shows the PixelReasoner prompt~\citep{wang2025pixel}, which adopts a specialized function-call format to enable the model to execute pixel-level image cropping operations based on normalized bounding boxes. Finally, Figure~\ref{fig:llm_judge_prompt} provides the VLM judge prompt used for automatic evaluation, where an expert system validates the correctness of the model's generated answer against the ground truth.
\input{figure/prompt.tex}

%% file: table/datasets.tex
\begin{table}[t]
\centering
\small
\begin{tabular}{lc}
\toprule
\textbf{Dataset} & \textbf{\#Queries} \\
\midrule
SlideVQA & 2215 \\
ViDoSeek & 1142 \\
MMLongBench-Doc & 859 \\
\bottomrule
\end{tabular}
\caption{Dataset statistics.}
\label{tab:dataset_stats}
\end{table}

%% file: figure/data_filter.tex
\begin{figure}[t]
    \centering
    \includegraphics[width=\linewidth]{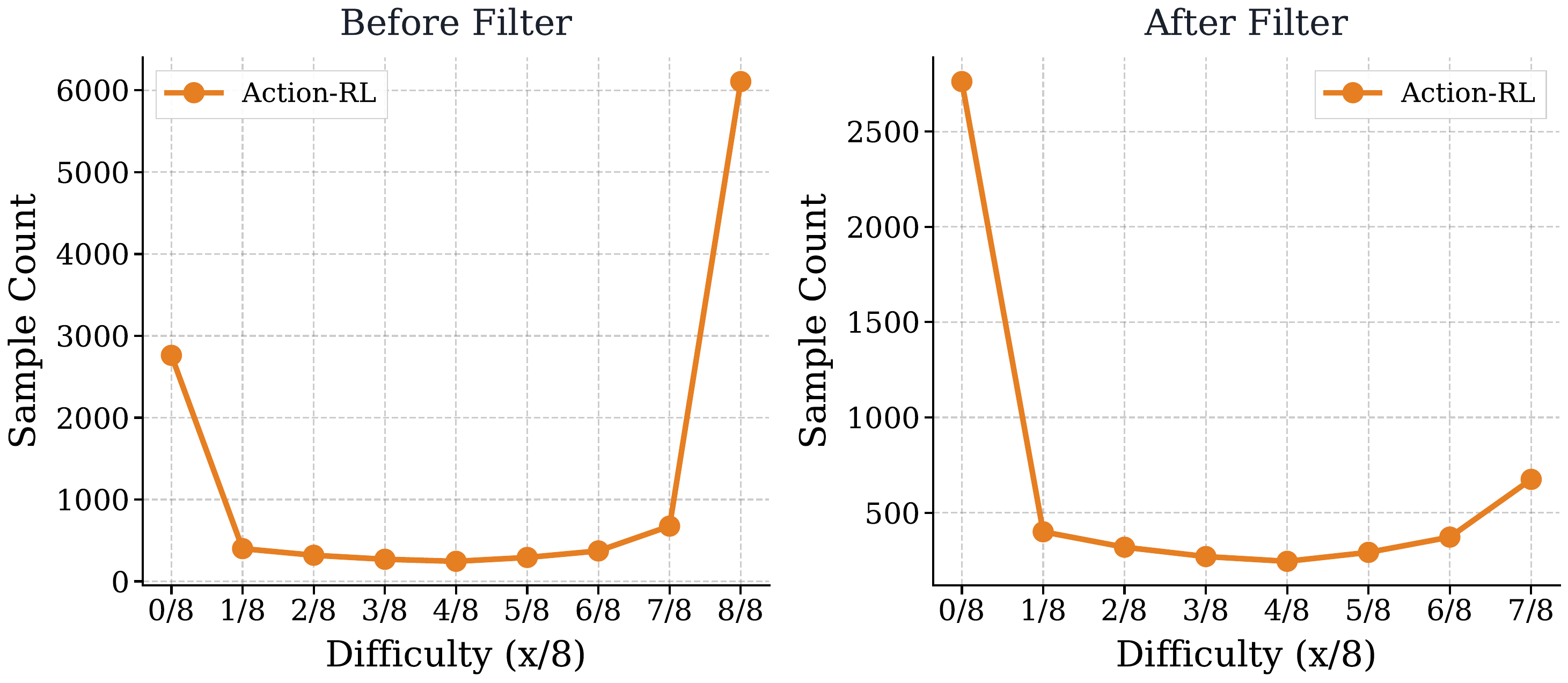}
    \caption{Distribution of sample difficulty before and after filtering in the Tool-Based Optimization training.}
    \label{fig:difficulty}
\end{figure}

%% file: table/stage1-hyperparameters.tex
\begin{table}[t]
\centering

\begin{tabular}{l c}
\hline
Epochs              & 1        \\
Rollout batch size  & 48       \\
Global batch size   & 8        \\
Max grad norm       & 1.0      \\
Data type           & bf16     \\
Learning rate       & $1.0\mathrm{e}{-6}$ \\
Weight decay        & $1.0\mathrm{e}{-2}$ \\
KL coefficient      & $1.0\mathrm{e}{-2}$\\
Rollout temperature & 1.0      \\
Epsilon             & 0.2      \\
Epsilon High        & 0.28     \\
Max prompt length   & 8192     \\
Max response length & 2048     \\
Image max pixels    & 401408 \\
\hline
\end{tabular}
\caption{Hyperparameters for Action RL.}
\label{tab:grpo-hparams}
\end{table}

%% file: table/stage2-hyperparameters.tex
\begin{table}[t]
\centering

\begin{tabular}{l c}
\hline
Epochs              & 3        \\
Rollout batch size  & 128       \\
Global batch size   & 64       \\
Max grad norm       & 1.0      \\
Data type           & bf16     \\
Learning rate       & $1.0\mathrm{e}{-6}$ \\
Weight decay        & $1.0\mathrm{e}{-2}$ \\
Rollout temperature & 1.0      \\
Epsilon             & 0.2      \\
Epsilon High        & 0.28     \\
Max prompt length   & 8192     \\
Max response length & 2048     \\
Image max pixels    & 401408 \\
\hline
\end{tabular}
\caption{Hyperparameters for Tool-based RL.}
\label{tab:grpo-hparams2}
\end{table}

%% file: table/retrieval.tex
\begin{table}[t]

\centering

\resizebox{\columnwidth}{!}{%

    \begin{tabular}{lcccc}

    \toprule

    \textbf{Dataset} & \textbf{Recall@1} & \textbf{Recall@3} & \textbf{Recall@5} & \textbf{MRR@5}  \\

    \midrule

    ViDoSeek        & 75.40 & 89.70 & 95.10 & 83.30 \\

    SlideVQA        & 92.91 & 98.10 & 98.96 & 95.53 \\

    MMLongBench-Doc & 49.00 & 61.70 & 66.50 & 55.70 \\

    \midrule

    \textbf{Avg.}   & 72.44 & 83.17 & 86.85 & 78.18 \\

    \bottomrule

    \end{tabular}%

}
\caption{Retrieval performance of ColQwen2 on three datasets. Metrics are Recall@k and MRR@5 (\%).}

\label{tab:colqwen2_retrieval_all}
\end{table}

%% file: table/acc.tex
\begin{table*}[t]
\centering
% \small
% \setlength{\tabcolsep}{3.2pt}
\resizebox{1.0\textwidth}{!}
{\begin{tabular}{lccccccccccccc}
\toprule
\multirow{3}{*}{\textbf{Methods}}
& \multicolumn{3}{c}{\textbf{In Domain}} 
& \multicolumn{9}{c}{\textbf{Out of Domain}} 
& \multirow{3}{*}{\textbf{Avg.}} \\
\cmidrule(lr){2-4}\cmidrule(lr){5-13} 
& \multicolumn{3}{c}{\textbf{SlideVQA}} 
& \multicolumn{6}{c}{\textbf{MMLongBench-Doc}} 
& \multicolumn{3}{c}{\textbf{ViDoSeek}} 
& \\
\cmidrule(lr){2-4}\cmidrule(lr){5-10}\cmidrule(lr){11-13} 
& Single & Multi & Overall 
& TXT & TAB & CHA & FIG & LAY & Overall 
& Single & Multi & Overall 
& \\
\midrule

% ===== Base models =====
\rowcolor{gray!20}\multicolumn{14}{l}{\textbf{\textit{Prompting Methods}}} \\\midrule
Direct
& 59.16 &49.74  &56.75 
& 35.91 &25.35  &33.15  &29.90  &30.51  &32.36 
& 22.64 &55.33  &36.87 
& 41.96 \\
TOT
& 55.04 &35.45  &50.02 
& 34.23 &23.96  &30.90  &33.89  &36.44  &33.88 
& 21.24 &59.15  &37.74 
&  40.56 \\
GOT
& 60.92 &51.32  &58.47 
& 37.58 &25.81  &32.02  &33.89  &35.59  &35.04 
&  24.50 &55.13  & 37.83
& 43.78 \\\midrule

% ===== VLRMs =====
\rowcolor{gray!20}\multicolumn{14}{l}{\textbf{\textit{Vision-Language Reasoning Models (VLRMs)}}} \\\midrule
R1-Onevision
& 59.89 &47.80  &56.79 
& 37.58 &27.65  &32.58  &36.88  &33.05  &36.09 
& 16.43 &57.95  & 34.50
&  42.46 \\
Vision-R1 
& 72.63 &\underline{56.61}  &\underline{68.53} 
& 41.28 &28.11  &\underline{38.20}  &\underline{40.20}  &\textbf{42.37}  &\underline{40.28} 
& 32.71 &60.56  &44.83 
& 51.21  \\
ThinkLite-VL
&65.84  &53.97  & 62.80
& 37.25 &27.65  &30.34  &37.21  &33.05  &35.74 
& 26.36 &61.77  &41.77 
& 46.77 \\
OpenVLThinker
& 69.24 &52.38  &64.92 
& 41.61 &\underline{28.57}  &33.71  &35.88  &\underline{38.14}  &37.49 
&29.77  &61.97  &43.78
& 48.73 \\
VisionMatters
& 61.47 &51.85  & 59.01
& 37.25 &24.88  &33.15  &33.22  &33.90  &34.58 
& 24.34 &61.37  &40.46 
&  44.68\\\midrule

% ===== MRAGs =====
\rowcolor{gray!20}\multicolumn{14}{l}{\textbf{\textit{Multimodal Retrieval-Augmented Generation Models (MRAGs)}}} \\\midrule
VisDom
& 58.92 &49.38  &56.48 
& 29.87 &20.28  &24.72  &21.59  &27.12  &26.78 
& 18.76 &57.14  &35.46 
&  39.57 \\
MM-Search-R1
& 65.47 &50.09  &61.53 
& 35.57 &23.96  &34.27  &33.89  &33.05  &33.99 
& 25.12 &61.37  & 40.89
& 45.47 \\
% R1-Router
% & 71.36 &59.44  & 68.31
% & 48.99 &35.48  &45.51  &42.86  &47.46  &46.68 
% & 29.61 &70.62  &47.46 
% & 54.15 \\
EVisRAG
& \underline{73.85} &52.20  &68.31 
&  37.92& 27.19 &32.02  &33.89  &\underline{38.14}  &35.86 
& 42.33 &61.67  &50.53 
& 51.56  \\\midrule

% ===== Ours =====
\rowcolor{gray!20}\multicolumn{14}{l}{\textbf{\textit{Tool-Enhanced VLMs}}} \\\midrule
Pixel-Reasoner
& 70.33 & 54.14 &66.19
&\underline{42.09} &26.73 &36.52 &\underline{40.20} &\underline{38.14} &39.28
& 36.55 &63.12 &48.51
& 51.32 \\
VRAG-RL
& 64.93 &54.67  &62.30 
& 36.58 &\textbf{33.90}  &\textbf{39.33}  &25.81  &33.55  &35.97 
& 24.81 &61.57  &40.81 
& 46.36 \\
\textbf{\method{}}
& \textbf{76.78} & \textbf{61.73} & \textbf{72.78} 
& \textbf{43.96} & 27.19 & 36.52 & \textbf{40.86} & \underline{38.14} & \textbf{40.51}
& \textbf{44.34} & \textbf{65.39} & \textbf{53.50}
& \textbf{55.60} 
\\
\bottomrule
\end{tabular}}
\caption{
Overall performance by using accuracy score for evaluation. The best results are highlighted in \textbf{bold}, and the second-best are \underline{underlined}.
}
\label{tab:main_overall_acc}
\end{table*}

%% file: table/Tool_Statistics.tex
\begin{table}[t]
    \centering
    \small

    \begin{tabular}{llr} 
        \toprule
        \textbf{Tool Name} & \textbf{Count} & \textbf{Frequency} \\
        \midrule
        \texttt{read\_text\_element} & 964 & 64.26\%\\
        \texttt{read\_numeric\_value} & 626 & 41.73\% \\
        \texttt{identify\_entity\_attribute} & 259 & 17.26\%\\
        \texttt{compare\_values} & 259 & 17.26\% \\
        \texttt{locate\_visual\_element} & 245 & 16.33\% \\
        \texttt{compute\_percentage} & 189 & 12.60\% \\
        \texttt{infer\_missing\_information} & 41 & 2.73\% \\
        \texttt{subtract\_values} & 20 & 1.33\% \\
        \texttt{add\_values} & 5 & 0.33\%\\
        \texttt{count\_matching\_values} & 3 & 0.20\% \\
        \bottomrule
    \end{tabular}
        \caption{Frequency of tool usage sorted by count.}
    \label{tab:tool_frequency}
\end{table}

%% file: figure/cot_analyse.tex
\begin{figure*}[t]
    \centering
    \begin{subfigure}[t]{0.32\linewidth}
        \centering
        \includegraphics[width=\linewidth]{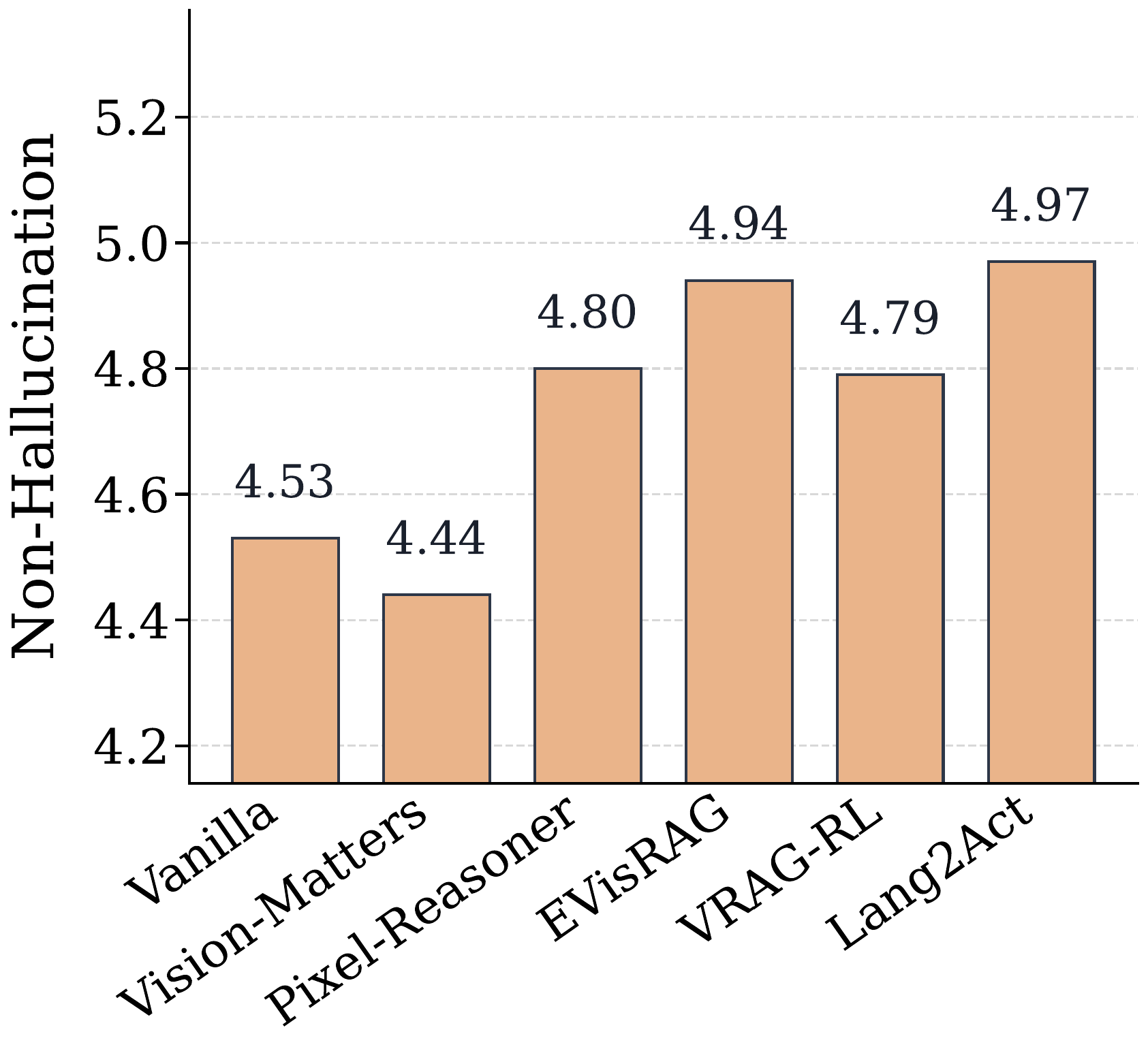}
        \caption{Non-hallucination performance.}
        \label{fig:non_hallucination}
    \end{subfigure}
    \hfill
    \begin{subfigure}[t]{0.32\linewidth}
        \centering
        \includegraphics[width=\linewidth]{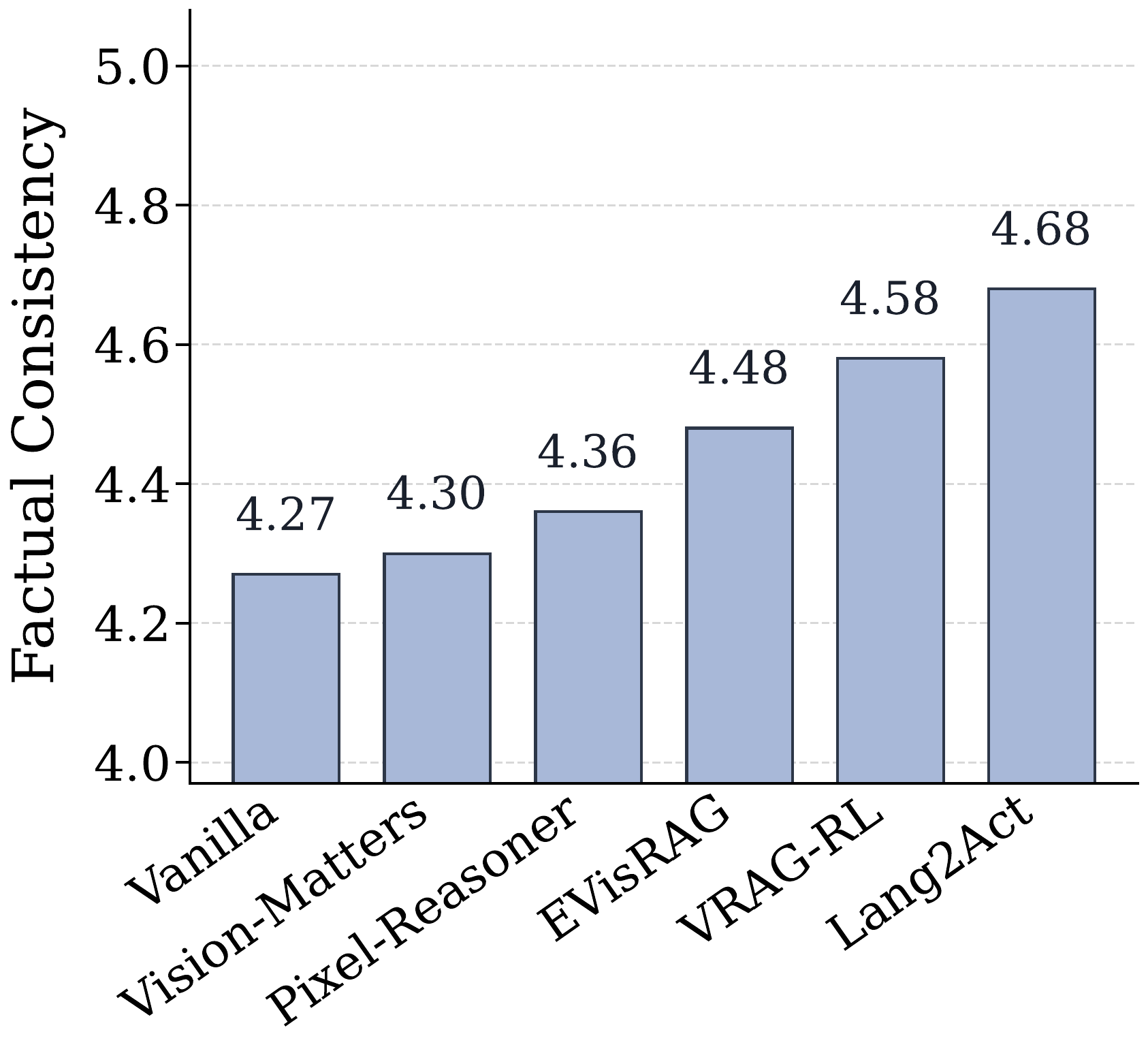}
        \caption{Factual consistency scores.}
        \label{fig:factual_consistency}
    \end{subfigure}
    \hfill
    \begin{subfigure}[t]{0.32\linewidth}
        \centering
        \includegraphics[width=\linewidth]{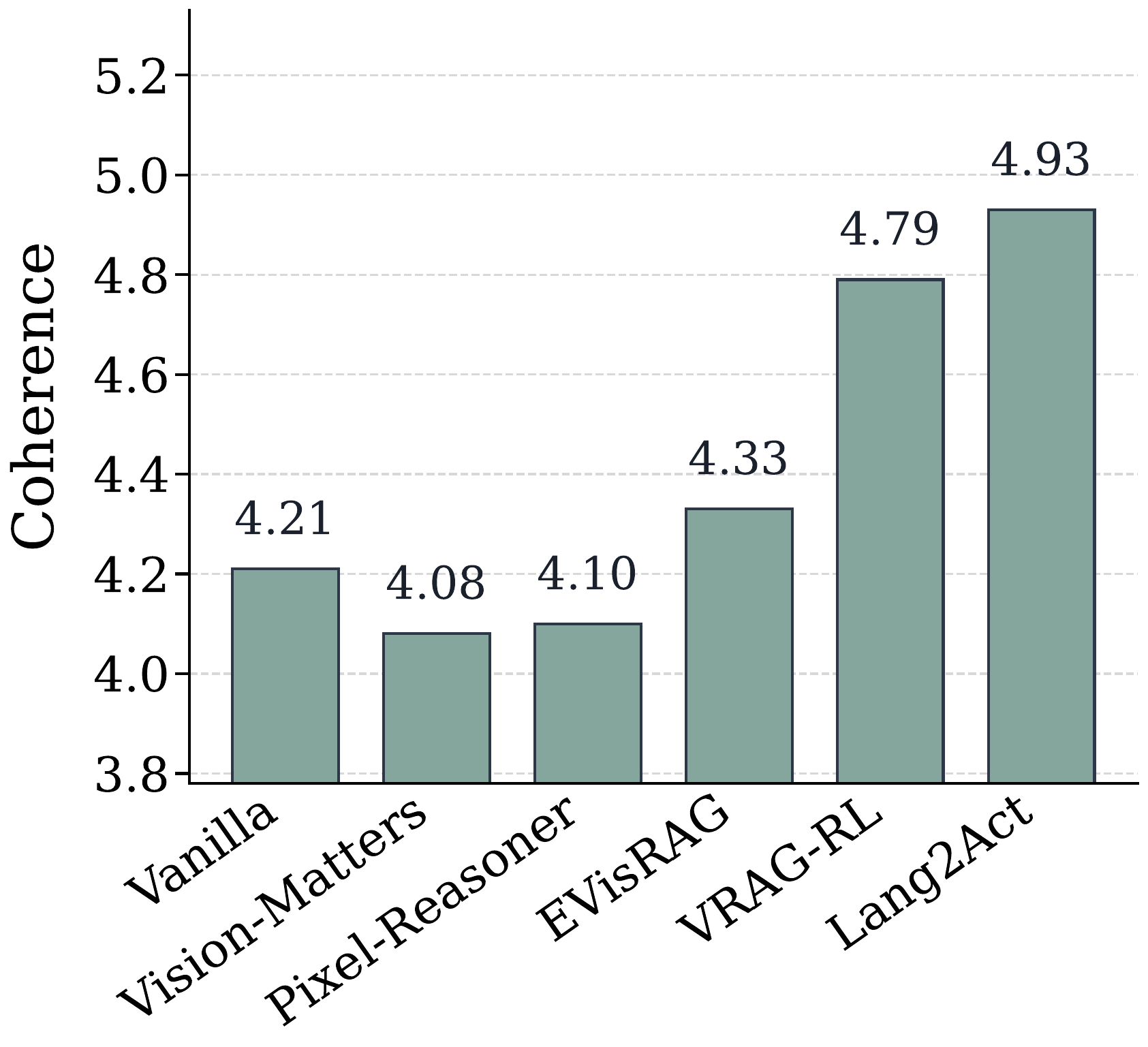}
        \caption{Coherence evaluation.}
        \label{fig:coherence}
    \end{subfigure}

    \caption{
Large language model–based evaluation of response quality under successful attention perception.
All scores are computed only on samples where the model attention successfully perceives the relevant (golden) regions.
We report performance from three complementary perspectives:
(a) non-hallucination,
(b) factual consistency, and
(c) coherence,
to assess the quality of model responses under reliable visual perception.
}
    \label{fig:quality_comparison}
\end{figure*}

%% file: table/analyse1.tex
\begin{table}[t]
\centering
\resizebox{\linewidth}{!}{
\begin{tabular}{lcccc}
\toprule
\textbf{Methods} & \textbf{Vidoseek} & \textbf{SlideVQA} & \textbf{MMBench} & \textbf{Avg.} \\
\hline
Vanilla         & 72.85 & 64.60 & 32.56 & 56.67 \\
R1-OneVision     & 73.56 & 67.67 & 36.05 & 59.09 \\
Vision-R1       & 71.28 & 72.19 & 38.89  & 60.78 \\
ThinkLite-VL    & 75.48 & 71.83 & 39.02 & 62.11 \\
OpenVLThinker   & 70.23 & 72.91 & 38.37 &60.26 \\
VisionMaters    & 73.29 & 70.56 & 37.86 &  60.57 \\
Pixel-Reasoner  & 71.54 & 69.53 & 38.16 & 59.74 \\
MM-Search-R1    & 77.58 & 71.69 & 37.98 & 62.41 \\
EVisRAG         & 72.94 & 74.63 & 40.83 & 62.80\\
VRAG-RL         & 73.73 & 75.67 & 39.31 & 62.90 \\
\textbf{\method{}}   & \textbf{79.95} & \textbf{79.68} & \textbf{43.28}& \textbf{67.63} \\
\bottomrule
\end{tabular}
}
\caption{Reasoning performance across three benchmarks under oracle retrieval.}
\label{tab:performance_comparison}
\end{table}

%% file: table/latency.tex
\begin{table}[h]
\centering
\begin{tabular}{l c}
\toprule
\textbf{Method} & \textbf{Latency (s)} \\
\hline
Vanilla & 48.77 \\
VisionMatters & 51.66 \\
OpenVLThinker & 72.93 \\
EVisRAG & 77.45 \\
Pixel-Reasoner & 94.45  \\
VRAG-RL & 123.78\\
Lang2Act & 50.49 \\
\bottomrule
\end{tabular}
\caption{Average end-to-end inference latency on SlideVQA}
\label{tab:latency_comparison}
\end{table}

%% file: figure/prompt.tex
% --- Vanilla RAG ---
% \begin{figure}[t] 
% \centering
% \begin{tcolorbox}[promptbox, title=Vanilla RAG Prompt.]
% \textbf{System Prompt:} \\[2pt]
% Answer the given question. You must conduct reasoning inside {<think>} and {</think>} first every time you get new information.
% After reasoning, you should directly provide the answer inside {<answer>} and {</answer>}, without detailed explanations.
% For example, {<answer> Beijing </answer>}. \\[4pt]

% \tcblower
% \textbf{User Prompt:} \\[2pt]
% Query: {\{Query Description\}} \\[2pt]
% Images: {\{Retrieved Images\}}
% \end{tcolorbox}
% \caption{Prompt of Vanilla RAG.}
% \label{fig:vanilla_rag_prompt}
% \end{figure}

% --- Simple-CoT (replace Vanilla RAG) ---
% \begin{figure}[t] 
% \centering
% \begin{tcolorbox}[promptbox, title=Vanilla RAG Prompt.]
% \textbf{System Prompt:} \\[2pt]
% You are a helpful AI assistant tasked with answering a question based on \{{num\_images}\} provided image(s). \\[6pt]

% Your response should follow this  format, with no text outside the tags: \\
% {<think>...</think>} \\
% {<answer>...</answer>} \\[6pt]

% Guidance: \\
% 1.\quad In {<think>}, provide your step-by-step reasoning for analyzing the images and deriving the answer. \\
% 2.\quad In {<answer>}, provide only the final, concise answer. \\[4pt]

% \tcblower
% \textbf{User Prompt:} \\[2pt]
% Query: {\{Query Description\}} \\[2pt]
% Images: {\{Retrieved Images\}}
% \end{tcolorbox}
% \caption{Prompt of Vanilla RAG.}
% \label{fig:simple_cot_prompt}
% \end{figure}

% --- Vanilla Prompt ---
\begin{figure*}[t] 
\centering
\begin{tcolorbox}[promptbox, title=Vanilla Prompt.]
\textbf{System Prompt:} \\[2pt]
Answer the given question based on the \{{num\_images}\} image(s) provided. 
You must conduct reasoning inside {<think>} and {</think>} first. 
After reasoning, you should directly provide the answer inside {<answer>} and {</answer>}, 
without detailed illustrations. \\[6pt]

\tcblower
\textbf{User Prompt:} \\[2pt]
Query: {\{Query Description\}} \\[2pt]
Images: {\{Retrieved Images\}}
\end{tcolorbox}
\caption{Prompt of Vanilla.}
\label{fig:vanilla2_prompt}
\end{figure*}

% meta-prompt
% --- XML-CoT VQA ---
\begin{figure*}[t]
\centering
\begin{tcolorbox}[promptbox, title=Action RL Prompt.]
\textbf{System Prompt:} \\[2pt]
You are a specialized AI assistant for visual question answering based on multiple provided document images. 
Your task is to answer the user's question by carefully analyzing all images. \\[6pt]

Your response must strictly follow this format: \\
{<think>...</think>} \\
{<description>...</description>} \\
{<answer>...</answer>} \\[6pt]

Guidance: \\
- You have exactly \{{num\_images}\} image(s). Analyze each briefly in {<think>}, then conclude which one(s) you used. \\
- In {<description>}, describe only the visual evidence you actually used, and clearly indicate where it appears in the image. \\
- In {<answer>}, output only the final concise answer. \\[4pt]

\tcblower
\textbf{User Prompt:} \\[2pt]
Query: {\{User Question\}} \\[2pt]
Images: {\{Retrieved Images\}}
\end{tcolorbox}
\caption{Prompt of  Action RL.}
\label{fig:xml_cot_vqa_prompt}
\end{figure*}

\begin{figure*}[t]
\centering
\begin{tcolorbox}[promptbox, title=Tools Curation Prompt.]
% 保持正常字号，去除了内容中的加粗
\textbf{System Prompt:} \\[2pt]
You are the Lead Architect of a Document Visual Reasoning System. Deconstruct questions into atomic, structure-aware cognitive operations. \\[3pt]

\#\#\# CORE PHILOSOPHY \\
1) Structure Awareness: Tools must reflect layout (rows, columns, axes). \\
2) Atomic Data Extraction: Locate region first, then extract data. \\
3) Analytical Calculation: Define precise math tools (subtract, rank\_values). \\[3pt]

\#\#\# CURRENT TOOL POOL \\
\{tool\_pool\_text\} \\[3pt]

\#\#\# GUIDELINES: DESIGNING DOCUMENT TOOLS \\
- Tables/Grids: Navigate rows/columns (e.g., locate\_table\_row, read\_cell\_value). \\
- Charts/Graphs: Map visuals to values (e.g., map\_bar\_to\_axis). \\
- Reasoning: Define specific logic tools for calculation/comparison. \\[3pt]

\#\#\# COGNITIVE CAPABILITY SPECTRUM \\
* Layout: (locate\_row/col, find\_title, find\_legend, intersect\_regions) \\
* Data: (read\_text, read\_numeric, extract\_key\_value\_pair) \\
* Chart: (trace\_line\_trend, get\_bar\_height, map\_color\_to\_category) \\
* Math \& Logic: (compute\_pct, subtract\_values, find\_max, count\_rows) \\
* Verify: (verify\_signature\_presence, check\_checkbox\_status) \\[3pt]

\#\#\# OUTPUT FORMAT (MANDATORY) \\
1. New Definitions: \texttt{DEFINE\_TOOL: name || args || desc} \\
2. Applications: \texttt{\textless tool name="..." args="..."\textgreater reasoning\textless /tool\textgreater} \\
3. End: \texttt{END\_OF\_TOOLS} \\[3pt]

\#\#\# EXAMPLE (Structure \& Math) \\
DESC: Found 'Q3 Revenue', read value, compared to 'Q2', calculated growth. \\
OUTPUT: \\
DEFINE\_TOOL: subtract\_values || val1, val2 || Calculate difference. \\
\textless tool name="locate\_table\_row" args="row 'Q3 Revenue'"\textgreater Row 4\textless /tool\textgreater \\
\textless tool name="read\_cell\_value" args="Row 4, col 'Amount'"\textgreater\$150M\textless /tool\textgreater \\
\textless tool name="subtract\_values" args="150, 100"\textgreater50\textless /tool\textgreater \\
END\_OF\_TOOLS

\tcblower
\textbf{User Prompt:} \\[2pt]
Analyze the reasoning steps. \\[2pt]
DESCRIPTION: \{description\} \\[2pt]
OUTPUT:
\end{tcolorbox}
\caption{Prompt of Tools Curation.}
\label{fig:tools_curation_prompt}
\end{figure*}

% --- Lang2ACT---
\begin{figure*}[t]
\centering
\begin{tcolorbox}[promptbox, title=Lang2Act Prompt.]
\textbf{System Prompt:} \\[2pt]
You are a specialized AI assistant for visual question answering. 
Your task is to answer the user's question by carefully analyzing all the provided images. \\[6pt]

Your response must strictly follow this XML format: \\
{<think>...</think>} \\
{<description>...</description>} \\
{<answer>...</answer>} \\[6pt]
Guidance: \\
1.\quad In {<think>}, analyze all {\{num\_images\}} images and state which one(s) contain relevant evidence. \\
2.\quad In {<description>}, focus only on the selected images and describe your reasoning process using the tools below. \\
3.\quad In {<answer>}, provide only the final, concise answer grounded in visual evidence. \\[4pt]
Available Tools for {<description>}: \\[-2pt]
\quad-- {<tool name="locate\_visual\_element" args="Image k: structural hint">} Locate specific visual elements or regions based on structural hints. {</tool>} \\
\quad-- {<tool name="read\_text\_element" args="Image k: locator/region">} Read and transcribe visible text from the located region. {</tool>} \\
\quad-- {<tool name="read\_numeric\_value" args="Image k: data point">} Extract specific numeric values or counts from visual elements. {</tool>} \\
\quad-- {<tool name="identify\_entity\_attribute" args="Image k: entity">} Identify specific attributes associated with entities. {</tool>} \\
\quad-- {<tool name="compare\_values" args="Image k: value A vs value B">} Compare quantitative values to determine ordering or equality. {</tool>} \\
\quad-- {<tool name="compute\_percentage" args="part\_value, total\_value">} Compute the percentage based on given values. {</tool>} \\
\quad-- {<tool name="infer\_missing\_information" args="Image k: data">} Infer missing information based on given data. {</tool>} \\[6pt]

\tcblower
\textbf{User Prompt:} \\[2pt]
Query: {\{User Question\}} \\[2pt]
Images: {\{Retrieved Images\}}
\end{tcolorbox}
\caption{
Prompt of Lang2Act.
}
\label{fig:lexicon_prompt}
\end{figure*}

% --- EviSRAG Prompt ---
\begin{figure*}[t]
\centering
\begin{tcolorbox}[promptbox, title=EVisRAG Prompt.]
\textbf{System Prompt:} \\[2pt]
You are an AI Visual QA assistant. I will provide you with a question and several images. Please follow the four steps below. \\[6pt]

\textbf{Step 1: Observe the Images} \\
First, analyze the question and consider what types of images may contain relevant information. Then, examine each image one by one, paying special attention to aspects related to the question. Identify whether each image contains any potentially relevant information. \\
Wrap your observations within {<observe>...</observe>} tags. \\[6pt]

\textbf{Step 2: Record Evidences from Images} \\
After reviewing all images, record the evidence you find for each image within {<evidence>...</evidence>} tags. \\
If you are certain that an image contains no relevant information, record it as: {[i]: no relevant information} (where {i} denotes the index of the image). \\
If an image contains relevant evidence, record it as: {[j]: [the evidence you find for the question]} (where {j} is the index of the image). \\[6pt]

\textbf{Step 3: Reason Based on the Question and Evidence} \\
Based on the recorded evidence, reason about the answer to the question. \\
Include your step-by-step reasoning within {<think>...</think>} tags. \\[6pt]

\textbf{Step 4: Answer the Question} \\
Provide your final answer based only on the evidence you found in the images. \\
Wrap your answer within {<answer>...</answer>} tags. \\
Avoid adding unnecessary contents in your final answer; for example, if the question is a yes/no question, simply answer {<answer>yes</answer>} or {<answer>no</answer>}. \\
If none of the images contain sufficient information to answer the question, respond with {<answer>insufficient to answer</answer>}. \\[6pt]

\textbf{Formatting Requirements:} \\
Use the exact tags {<observe>}, {<evidence>}, {<think>}, and {<answer>} for structured output. \\
It is possible that none, one, or several images contain relevant evidence. \\
If you find no evidence or too little evidence to answer the question, follow the instructions above for insufficient information. \\[4pt]

\tcblower
\textbf{User Prompt:} \\[2pt]
Query: {\{User Question\}} \\[2pt]
Images: {\{Retrieved Images\}}
\end{tcolorbox}
\caption{Prompt of EVisRAG for evidence-structured visual question answering.}
\label{fig:evisrag_prompt}
\end{figure*}

% --- ToT ---
% \begin{figure}[t]
% \centering
% \begin{tcolorbox}[promptbox, title=ToT Prompt.]
% \textbf{System Prompt:} \\[2pt]
% You are an AI assistant. I will provide a query and \{{num\_images}\} image(s). You must use a 'Tree of Thoughts' approach to arrive at the answer. \\[6pt]
% Your response must strictly follow this format: \\
% {<think>...</think>} \\
% {<answer>...</answer>} \\[6pt]
% Guidance: \\
% In the first step (within the {<think>} tag): \\
% 1.\quad \textbf{Deconstruct the Problem}: Break down the main question into smaller, manageable sub-problems. \\
% 2.\quad \textbf{Generate Multiple Thoughts}: For each sub-problem, generate at least two potential lines of reasoning or "thoughts" on how to solve it using the provided images. \\
% 3.\quad \textbf{Evaluate Thoughts}: Assess each thought's validity. Analyze the evidence in the images that supports or refutes each thought. State which thoughts are dead ends and which are promising. \\
% 4.\quad \textbf{Conclude}: Based on your evaluation, synthesize the most promising thoughts to form a final, coherent reasoning chain. \\[4pt]
% In the second step (within the {<answer>} tag): \\
% Provide only the final, concise answer that results from your "Tree of Thoughts" analysis. If the question asks for a "yes" or "no", only provide that.

% \tcblower
% \textbf{User Prompt:} \\[2pt]
% Query: {\{User Question\}} \\[2pt]
% Images: {\{Retrieved Images\}}
% \end{tcolorbox}
% \caption{Prompt of Tree-of-Thoughts (ToT).}
% \label{fig:tot_prompt}
% \end{figure}

% --- ToT ---
\begin{figure*}[t]
\centering
\begin{tcolorbox}[promptbox, title=ToT Prompt.]
\textbf{System Prompt:} \\[2pt]
You are an AI assistant. I will provide a query and \{{num\_images}\} image(s). You must use a 'Tree of Thoughts' approach to arrive at the answer. \\[6pt]

Follow these two steps: \\[4pt]

In the first step (within the {<think>} tag): \\
1.\quad \textbf{Deconstruct the Problem}: Break down the main question into smaller, manageable sub-problems. \\
2.\quad \textbf{Generate Multiple Thoughts}: For each sub-problem, generate at least two potential lines of reasoning or 'thoughts' on how to solve it using the provided images. \\
3.\quad \textbf{Evaluate Thoughts}: Assess each thought's validity. Analyze the evidence in the images that supports or refutes each thought. State which thoughts are dead ends and which are promising. \\
4.\quad \textbf{Conclude}: Based on your evaluation, synthesize the most promising thoughts to form a final, coherent reasoning chain. \\[6pt]

In the second step (within the {<answer>} tag): \\
Provide only the final, concise answer that results from your 'Tree of Thoughts' analysis. If the question asks for a 'yes' or 'no', only provide that.

\tcblower
\textbf{User Prompt:} \\[2pt]
Query: {\{User Question\}} \\[2pt]
Images: {\{Retrieved Images\}}
\end{tcolorbox}
\caption{Prompt of Tree-of-Thoughts (ToT).}
\label{fig:tot_prompt}
\end{figure*}

% --- GOT ---
% \begin{figure}[t]
% \centering
% \begin{tcolorbox}[promptbox, title=GOT Prompt.]
% \textbf{System Prompt:} \\[2pt]
% You are an AI assistant. I will provide a query and \{{num\_images}\} image(s). You must use a 'Graph of Thoughts' approach to solve the problem. \\[6pt]
% Your response must strictly follow this format: \\
% {<think>...</think>} \\
% {<answer>...</answer>} \\[6pt]
% Guidance: \\
% In the first step (within the {<think>} tag): \\
% 1.\quad \textbf{Generate Initial Thoughts}: Create several initial, independent thoughts or approaches to answering the question based on the images. \\
% 2.\quad \textbf{Transform and Refine}: For each thought, consider how it can be improved, refined, or combined with others. Merge promising thoughts into a synthesized line of reasoning and discard incorrect ones. \\
% 3.\quad \textbf{Structure as a Graph}: Explain your final reasoning process as a graph where thoughts are nodes and edges indicate refinement/merging. Show the progression from initial thoughts to the final synthesized conclusion. \\[4pt]
% In the second step (within the {<answer>} tag): \\
% Provide only the final, concise answer derived from your "Graph of Thoughts" analysis.

% \tcblower
% \textbf{User Prompt:} \\[2pt]
% Query: {\{User Question\}} \\[2pt]
% Images: {\{Retrieved Images\}}
% \end{tcolorbox}
% \caption{Prompt of Graph-of-Thoughts (GOT).}
% \label{fig:got_prompt}
% \end{figure}

% --- GOT ---
\begin{figure*}[t]
\centering
\begin{tcolorbox}[promptbox, title=GOT Prompt.]
\textbf{System Prompt:} \\[2pt]
You are an AI assistant. I will provide a query and \{{num\_images}\} image(s). You must use a 'Graph of Thoughts' approach to solve the problem. \\[6pt]

Follow these two steps: \\[4pt]

In the first step (within the {<think>} tag): \\
1.\quad \textbf{Generate Initial Thoughts}: Create several initial, independent thoughts or approaches to answering the question based on the images. \\
2.\quad \textbf{Transform and Refine}: For each thought, consider how it can be improved, refined, or combined with others. Merge promising thoughts into a more powerful, synthesized line of reasoning. Discard thoughts that are incorrect. \\
3.\quad \textbf{Structure as a Graph}: Explain your final reasoning process as a graph where thoughts are nodes. Show how you progressed from initial thoughts to the final synthesized conclusion. \\[6pt]

In the second step (within the {<answer>} tag): \\
Provide only the final, concise answer derived from your 'Graph of Thoughts' analysis.

\tcblower
\textbf{User Prompt:} \\[2pt]
Query: {\{User Question\}} \\[2pt]
Images: {\{Retrieved Images\}}
\end{tcolorbox}
\caption{Prompt of Graph-of-Thoughts (GOT).}
\label{fig:got_prompt}
\end{figure*}

% --- CoV ---
% \begin{figure}[t]
% \centering
% \begin{tcolorbox}[promptbox, title=CoV Prompt.]
% \textbf{System Prompt:} \\[2pt]
% You are an AI assistant. I will provide a query and \{{num\_images}\} image(s). You must use a 'Chain of Verification' process to ensure your answer is accurate. \\[6pt]
% Your response must strictly follow this format: \\
% {<think>...</think>} \\
% {<answer>...</answer>} \\[6pt]
% Guidance: \\
% In the first step (within the {<think>} tag): \\
% 1.\quad \textbf{Baseline Response}: First, generate a direct, initial answer without explaining the reasoning. \\
% 2.\quad \textbf{Plan Verification}: Create a list of specific verification questions you need to check the correctness of your baseline response. These should be concrete, fact-based questions. \\
% 3.\quad \textbf{Execute Verification}: Answer each verification question one by one, citing evidence from the images (indicate where in the image the evidence appears). \\[4pt]
% In the second step (within the {<answer>} tag): \\
% Based on the results of your verification, provide the final, corrected answer. If verification shows the baseline was wrong, state the corrected answer; if it was correct, confirm it.

% \tcblower
% \textbf{User Prompt:} \\[2pt]
% Query: {\{User Question\}} \\[2pt]
% Images: {\{Retrieved Images\}}
% \end{tcolorbox}
% \caption{Prompt of Chain-of-Verification (CoV).}
% \label{fig:cov_prompt}
% \end{figure}

% --- VRAG-RL ---
\begin{figure*}[t]
\centering
\begin{tcolorbox}[promptbox, title=VRAG-RL Prompt.]
\textbf{System Prompt:} \\[2pt]
Answer the given question. You must conduct reasoning inside {<think>} and {</think>} first every time you get new information.
After reasoning, if you lack knowledge, you may call a search engine via {<search> query </search>}. 
When an image is retrieved, you may crop it using {<bbox>[x1, y1, x2, y2]</bbox>}. 
Repeat as needed. If no further knowledge is needed, provide the answer within {<answer>} and {</answer>}. 
For example, {<answer> Beijing </answer>}. \\[4pt]

\tcblower
\textbf{User Prompt:} \\[2pt]
Query: {\{Query Description\}}
\end{tcolorbox}
\caption{Prompt of VRAG-RL.}
\label{fig:vrag_rl_prompt}
\end{figure*}
% --- PixelReasoner ---
\begin{figure*}[t]
\centering
\begin{tcolorbox}[promptbox, title=PixelReasoner Prompt.]
\textbf{System Prompt:} \\[2pt]
You are a helpful assistant. You may call one or more functions to assist with the user query. 
You are provided with function signatures within {<tools>} XML tags (specifically {crop\_image\_normalized} to zoom in based on normalized bbox). 
For each function call, return a json object with function name and arguments within {<tool\_call>} XML tags: 
{<tool\_call> \{ "name": ..., "arguments": ... \} </tool\_call>}. \\[4pt]

\tcblower
\textbf{User Prompt:} \\[2pt]
Query: {\{Query Description\}} \\[4pt]
\textit{Guidelines:} Understand the given visual information and the user query. Determine if it is beneficial to employ the given visual operations (tools). 
For a video, we can look closer by {select\_frames}. For an image, we can look closer by {crop\_image\_normalized}. 
Reason with the visual information step by step, and put your final answer within {\textbackslash boxed\{\}}.
\end{tcolorbox}
\caption{Prompt of PixelReasoner.}
\label{fig:pixel_reasoner_prompt}
\end{figure*}

% --- LLM Judge ---
\begin{figure*}[t]
\centering
\begin{tcolorbox}[promptbox, title=VLM Judge Prompt.]
\textbf{System Prompt:} \\[2pt]
You are an expert evaluation system for a question answering chatbot.
You will be given one evaluation item. You will see a query, a reference answer, and a generated answer.
Your task is to evaluate the correctness of the generated answer.
Your response MUST be exactly one line, formatted as {<judge>True</judge>} if the generated answer is correct, or {<judge>False</judge>} otherwise.
Do not add any other text or explanations. \\[4pt]

\tcblower
\textbf{User Prompt:} \\[2pt]
Query: {\{Query\}} \\[2pt]
Reference Answer: {\{Reference Answer\}} \\[2pt]
Generated Answer: {\{Model Answer To Evaluate\}} 
\end{tcolorbox}
\caption{Prompt of the automatic judge for single-item evaluation.}
\label{fig:llm_judge_prompt}
\end{figure*}

% --- LLM Judge for Coherence & Factuality ---
\begin{figure*}[t]
\centering
\begin{tcolorbox}[promptbox, title=LLM Judge Prompt for Reasoning Quality.]
\textbf{System Prompt:} \\[2pt]
You are an expert evaluator for Large Language Models. Your task is to evaluate the quality of a model's ``Chain of Thought'' (reasoning process) and final answer based on the user's Query and the provided Gold Answer.

Please evaluate the {[Model Response]} based on the following three specific dimensions. For each dimension, assign a score from 1 to 5 stars.

\textbf{1. Coherence (Reasoning Logic \& Fluency)} \\
\textit{Definition:} Evaluates whether the chain of thought is logically sound, structured, and easy to follow. \\
(1 Star: Disjointed/Confusing; 3 Stars: Readable but with leaps; 5 Stars: Perfectly smooth/Logical).

\textbf{2. Non-Hallucination (Faithfulness)} \\
\textit{Definition:} Evaluates whether the response contains fabricated information. \\
(1 Star: Major fabrications; 3 Stars: Minor errors; 5 Stars: Entirely truthful).

\textbf{3. Factual Consistency (Alignment with Gold Answer)} \\
\textit{Definition:} Evaluates whether the model's final conclusion aligns with the Gold Answer. \\
(1 Star: Contradictory; 3 Stars: Partially consistent; 5 Stars: Fully consistent).

\textbf{Output Format:} \\
Please strictly follow this format: \\
Coherence: [Score] \\
Non-Hallucination: [Score] \\
Factual Consistency: [Score] \\
Average: [Average Score] \\
Explanation: [Brief explanation]

\tcblower
\textbf{User Prompt:} \\[2pt]
\textbf{Query:} {\{Query\}} \\[2pt]
\textbf{Gold Answer:} {\{Gold Answer\}} \\[2pt]
\textbf{Model Response:} {\{Model Response\}}
\end{tcolorbox}
\caption{Prompt used for the automatic evaluation of reasoning coherence, faithfulness, and factual consistency.}
\label{fig:llm_judge_reasoning}
\end{figure*}